\documentclass[12pt]{article}

\usepackage{graphicx}   
\usepackage{subcaption} 
\usepackage{graphicx}   
\usepackage{makecell}   
\usepackage{booktabs}   
\usepackage{xcolor}     
\usepackage{pifont}     
\usepackage{array}      
\usepackage{enumitem}   
\usepackage{longtable}  
\usepackage{rotating}   
\usepackage{placeins}   
\usepackage{float}      
\usepackage{afterpage}  
\usepackage{rotfloat}   
\usepackage{tikz}       
\usepackage{authblk}    
\usepackage[american]{babel}
\usepackage{hyperref}
\usepackage{amsmath}
\usepackage{array}
\usepackage{caption} 
\usepackage{tabularx}
\usepackage{natbib}
\usepackage[a4paper, margin=1in]{geometry}
\usepackage{multirow}
\usepackage{multicol}
\usepackage{amssymb}
\usepackage{spverbatim}
\usepackage{listings}
\usepackage[frozencache=true,cachedir=.]{minted}
\setminted{
  breaklines=true,
  breaksymbolleft={},
  fontsize=\footnotesize,
  tabsize=4,
  obeytabs=true,
  baselinestretch=1,
}

\title{Exploring Big Five Personality and AI Capability Effects in LLM-Simulated Negotiation Dialogues}

\author[1,2]{\normalsize Myke C. Cohen}

\affil[1]{Aptima, Inc., MA, USA}

\affil[2]{Human Systems Engineering, Arizona State University, AZ, USA}

\author[3]{Zhe Su}

\affil[3]{Language Technologies Institute, Carnegie Mellon University, PA, USA}

\author[1]{Hsien-Te Kao}

\author[1]{Daniel Nguyen}

\author[1]{Spencer Lynch}

\author[3]{Maarten Sap}

\author[1]{Svitlana Volkova}

\begin{document}

\date{} 

\maketitle

\begin{abstract}
  This paper presents an evaluation framework for agentic AI systems in mission-critical negotiation contexts, addressing the need for AI agents that can adapt to diverse human operators and stakeholders. Using Sotopia as a simulation testbed, we present two experiments that systematically evaluated how personality traits and AI agent characteristics influence LLM-simulated social negotiation outcomes--a capability essential for a variety of applications involving cross-team coordination and civil-military interactions. Experiment 1 employs causal discovery methods to measure how personality traits impact price bargaining negotiations, through which we found that Agreeableness and Extraversion significantly affect believability, goal achievement, and knowledge acquisition outcomes. Sociocognitive lexical measures extracted from team communications detected fine-grained differences in agents' empathic communication, moral foundations, and opinion patterns, providing actionable insights for agentic AI systems that must operate reliably in high-stakes operational scenarios. Experiment 2 evaluates human-AI job negotiations by manipulating both simulated human personality and AI system characteristics, specifically transparency, competence, adaptability, demonstrating how AI agent trustworthiness impact mission effectiveness. These findings establish a repeatable evaluation methodology for experimenting with AI agent reliability across diverse operator personalities and human-agent team dynamics, directly supporting operational requirements for reliable AI systems. Our work advances the evaluation of agentic AI workflows by moving beyond standard performance metrics to incorporate social dynamics essential for mission success in human-centered defense operations.
\end{abstract}


\section{Introduction}

Large language models (LLMs) have advanced social simulations to unprecedented levels of fidelity. There is now a wide range of social interactions that can be simulated through LLM-driven agents, both interpersonal interactions (i.e., between people; \cite{zhouSOTOPIAInteractiveEvaluation2024}) and human-AI interactions \cite{volkovaVirTLabAugmentedIntelligenceinpress,nguyenExploratoryModelsHumanAI2024,cuiHumanAutonomyTeamingAutonomous2023,yangLLMBasedDigitalTwin2024}. In this study, we run LLM-driven negotiation simulations, which feature cooperative and competitive communication dynamics that must be balanced across social scenarios important for defense applications and beyond.

LLMs provide a novel framework for studying how negotiation unfolds and shapes social outcomes with respect to its various correlates. Among these are personality traits, which are factors of human variability that influence both cooperative \cite{bendellIndividualTeamProfiling2024} and competitive \cite{furnhamPersonalityInterpersonalInfluence2024} communication. Recent works suggest that large-scale LLM-driven simulations of social communication demonstrate qualities consistent with theoretical personality models, both in negotiation \cite{huangHowPersonalityTraits2024} and in human-AI team scenarios \cite{volkovaVirTLabAugmentedIntelligenceinpress}. In contrast, human subjects research methods are often limited in being able to investigate human variability factors like personality traits as controlled, independent experimental variables \cite{shadishExperimentalQuasiExperimentalDesigns2001a}. In this paper, we present two experiments leveraging Sotopia, an LLM-based simulation framework \cite{zhouSOTOPIAInteractiveEvaluation2024}, to investigate how Big Five personality traits and AI characteristics influence interpersonal and human-AI agent negotiation interactions. 

This work makes several novel contributions to the evaluation of agentic AI systems. First, we present the first systematic evaluation framework that explicitly examines the interplay between human personality traits (based on the Big Five model) and AI agent characteristics in negotiation contexts—a critical capability for mission-critical defense applications. While existing evaluation frameworks focus primarily on task completion metrics or tool usage accuracy, our approach uniquely captures the social dynamics essential for human-AI teaming effectiveness. Second, we employ causal discovery methods (CausalNex~\cite{beaumontCausalNex2021} and Causal Forest~\cite{atheyGeneralizedRandomForests2019b}) to quantify how personality traits causally impact negotiation outcomes, moving beyond correlational analyses typical in current agentic AI evaluations. This allows us to identify which trait combinations lead to optimal performance in high-stakes scenarios. Third, we introduce a comprehensive multi-dimensional evaluation methodology that combines: (1) scenario-based measures using Sotopia-Eval to assess goal achievement and interaction quality, (2) fine-grained lexical analytics to detect empathy patterns, moral foundations, and emotional dynamics that influence team trust and cooperation, and (3) post-interaction questionnaires that capture subjective evaluations critical for operational deployment. Finally, our dual-experiment design—examining both interpersonal (Experiment 1) and human-AI negotiations (Experiment 2)—provides actionable insights for designing AI agents that can adapt to diverse operator personalities and maintain performance under the stress and uncertainty characteristic of defense operations. This work establishes a repeatable methodology for stress-testing agentic AI reliability across the full spectrum of human variability, directly addressing the gap between laboratory AI performance and real-world operational requirements.

\section{Related Works}







\subsection{Personality \& Social Simulation}
Personality traits have long been defined relative to social communication processes and outcomes that can now be simulated through LLMs. For instance, the Big Five personality model \cite{mccraeIntroductionFiveFactorModel1992} can be traced back to early works investigating vocabulary words for describing oneself and others \cite{allport1936trait}, which \cite{cattell1946description} used to create rating scales comprising mostly adjectives about people's social qualities. These scale items were eventually refined and grouped together into the five factors \cite{fiske1949consistency, tupes1961recurrent, norman1963toward} that are now popularly known as the Big Five personality traits: Agreeableness, Conscientiousness, Extraversion, Neuroticism, and Openness. 

Partly due to its lexical origins, the Big Five model has been foundational in investigating the breadth of social behaviors that LLMs can simulate. Recent works increasingly leverage prompts derived from Big Five Inventory (BFI) questionnaire items (e.g., \cite{costa2008revised}) to define LLM personas in various social contexts, including collaborative writing \cite{frischLLMAgentsInteraction2024}, price bargaining \cite{huangHowPersonalityTraits2024}, and search-and-rescue team communication \cite{volkovaVirTLabAugmentedIntelligenceinpress, nguyenExploratoryModelsHumanAI2024}. Our present study similarly defines Big Five personality traits through BFI item prompts to elicit personality-driven differences in simulated negotiations.

In addition to simulating Big Five personality traits, BFI questionnaires have also inspired techniques to describe LLM personality traits \cite{karraEstimatingPersonalityWhiteBox2023, huangWhoChatGPTBenchmarking2024}. However, recent uses of such frameworks provide mixed evidence for the efficacy of BFI-based prompts in social simulations of \textit{human} personality traits. On one hand, prompting LLMs to adopt specific levels of Big Five personality traits consistently results in expected trait-associated social behaviors \cite{huangHowPersonalityTraits2024, jiangPersonaLLMInvestigatingAbility2024, duanPowerPersonalityHuman2025, frischLLMAgentsInteraction2024}. On the other hand, some findings suggest an incongruence between ostensibly personality-driven LLM behaviors versus LLM-generated self-descriptions \cite{aiSelfknowledgeActionConsistent2024} or responses to BFI psychometric instruments \cite{petrovLimitedAbilityLLMs2024, molchanovaExploringPotentialLarge2025}. To illustrate, a ``high Agreeableness'' LLM agent may express that it desires to maximize mutual outcomes during a negotiation, but subsequently average low on post-simulation BFI questionnaire responses, and vice versa. 


\subsection{Evaluating Simulated Negotiations}
We adopt Raiffa's \cite{raiffaArtScienceNegotiation1982} definition of negotiation as a structured or semi-structured interaction where two or more parties exchange bids with the goal of reaching mutual agreement on specific terms or resources. In social negotiation settings, achieving mutually-beneficial negotiation outcomes can be complicated by personality-linked emotions, moral stances, and perspective differences \cite{huangHowPersonalityTraits2024}. Indeed, decades of Big Five personality research demonstrates that certain personality traits can significantly impact negotiation outcomes \cite{gilkey1986role, sassPersonalityNegotiationPerformance2015, kangHowInterviewersRespond2015}, with traits like Extraversion and Agreeableness having positive or negative effects depending on the competitiveness of the negotiation setting \cite{barry1998bargainer, amanatullahNegotiatorsWhoGive2008}. This has also been demonstrated experimentally across numerous simulated contexts, including all-human and human-AI team operations of remotely-piloted aerial vehicles \cite{gormanTeamCoordinationDynamics2010, demirTeamCoordinationDynamics2017}; in price bargaining between a human buyer and an AI seller \cite{falcaoBigFivePersonality2018}; and, most recently, between two LLM agents \cite{huangHowPersonalityTraits2024}. As such, we selected negotiation outcomes as one of our primary measures for examining prompt-based LLM personality trait impacts in this study.

Beyond direct negotiation outcomes communication patterns provide a well-established window into subtle but impactful social, cognitive, and emotional processes linked to personality traits \citep{pennebakerLinguisticStylesLanguage1999, tausczikPsychologicalMeaningWords2010}. Lexical analyses have been extensively used to reveal how traits such as Agreeableness, Extraversion, or Neuroticism influence emotional expressiveness, prosociality, and interpersonal alignment in negotiation contexts \citep{barry1998bargainer, pennebakerLinguisticStylesLanguage1999}. For example, sentiment polarity measures provide direct insights into how personality traits shape affective language, which are crucial for both competitive and cooperative negotiations \citep{chawla2023towards}. Empathy indicators reflect personality-driven differences in how negotiators acknowledge and respond to their partner’s emotions and intents, influencing relational outcomes \citep{grazianoAgreeablenessEmpathyHelping2007, leeDoesGPT3Generate2022}. Analyses of nuanced indicators of a communicator's moral values and connotative perspectives reveal how personality traits guide implicit ethical considerations, social alignment, and subtle communicative strategies that can substantially influence negotiation trajectories and outcomes \citep{garten2016morality, rashkinConnotationFramesDataDriven2016a, grahamLiberalsConservativesRely2009}. Finally, indicators of subjectivity, toxicity, and hate speech helps account for hostile or antagonistic engagement tendencies \citep{hanuDetoxify2020, rashkinConnotationFramesDataDriven2016a}, which align with classical descriptors for low Agreeableness and high Neuroticism. We take these various lexical measures in concert as we consider personality-linked impacts on our simulated negotiation scenarios.

Alongside {\bf outcome-based} and {\bf lexical} measures, {\bf questionnaire} measures have also been used to measure humans' perceptions of their experiences during social negotiations. Research shows that subjective measures of a negotiation partner's trustworthiness, fairness, and reliability are also influenced by personality traits \cite{curhanWhatPeopleValue2006, barry1998bargainer, elfenbeinAreNegotiatorsBetter2008}, even when negotiation outcomes are held constant. For instance, negotiators high in Agreeableness tend to foster more positive impressions, while high Neuroticism is associated with increased post-negotiation frustration \cite{dimotakisMindHeartLiterally2012}. Similar findings have been observed in simulated human-agent interactions, demonstrating that users reliably detect and respond to personality cues in AI agents, with subjective evaluations providing evidence of whether those traits were effectively conveyed \cite{prajodEffectVirtualAgent2019, zhouTrustingVirtualAgents2019}. As such, we include post-interaction questionnaire items to capture how our simulations approximate human subjective evaluations of the negotiation experience, relative to personality traits and AI characteristics.

\section{Current Study}

\subsection{Simulation Framework}
\begin{figure*}[t]
    \centering
    \includegraphics[width=0.75\linewidth]{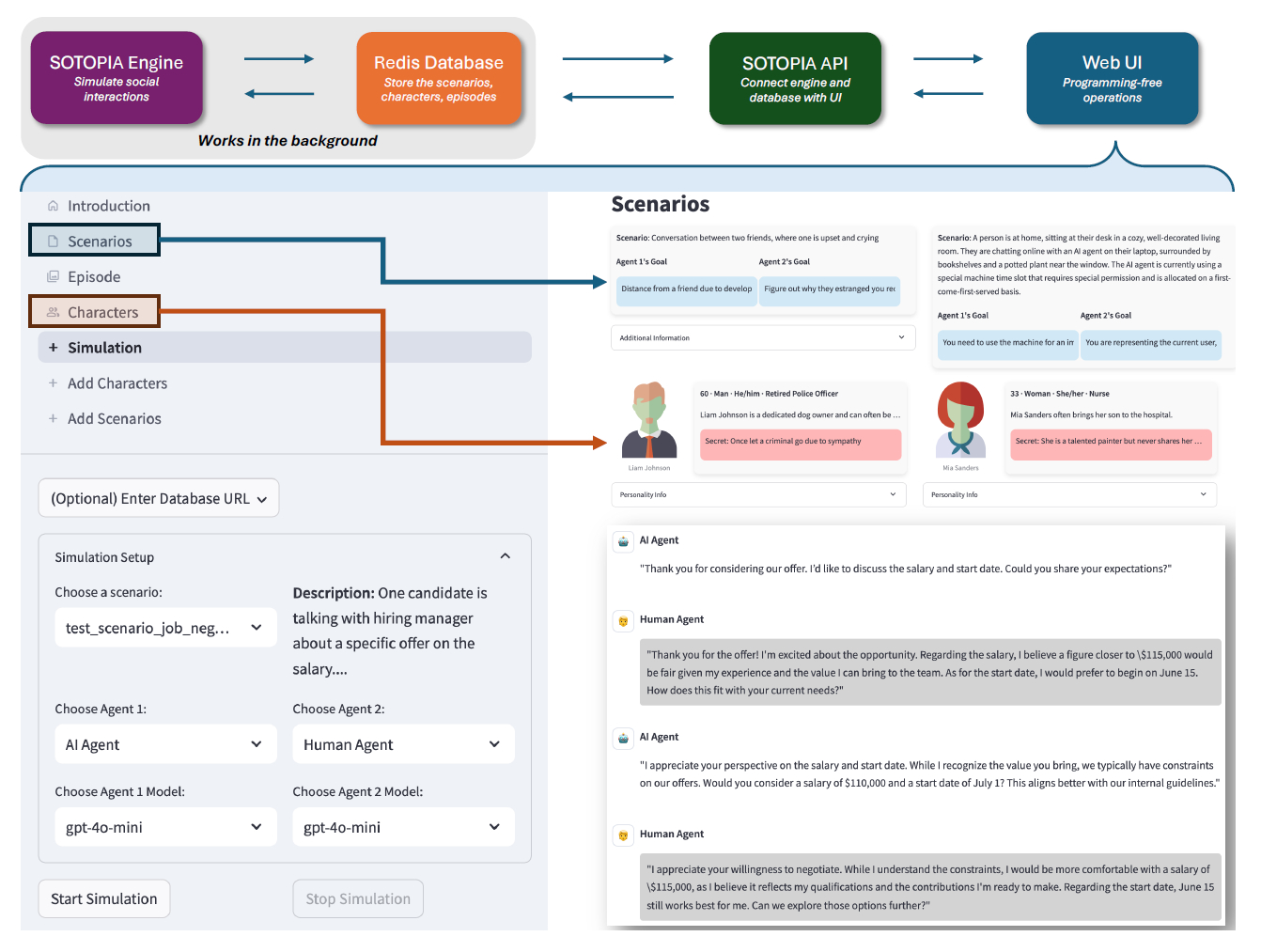}
    \caption{Sotopia simulation framework~\cite{Zhou2024SOTOPIA}.}
    \label{fig:sotopia-framework}
\end{figure*}

We used Sotopia \cite{zhouSOTOPIAInteractiveEvaluation2024} as our modeling framework for simulating various interpersonal scenarios between two agents, illustrated in Figure~\ref{fig:sotopia-framework}. Sotopia simulation episodes occur in dialogue form, in which agents took turns playing their assigned characters while dynamically interacting to achieve their objectives. Three main parameters are specified to generate a Sotopia episode: (1) a scenario; (2) character profiles; and (3) characters' respective social goals. Our study adds a fourth simulation parameter---AI characteristics---to investigate the concurrent impacts of AI characteristics and simulated personality traits in human-AI simulations.

\subsubsection{Scenario} Sotopia scenarios comprise shared information (e.g., location, time, situation) that provides overall context for individual agent-specific goals to guide their behavior. An example is shown in Figure~\ref{fig:sotopia-framework}, where the scenario is described as  "one candidate is talking with the hiring manager...". Scenarios can also include constraints to other parameter attributes e.g., only valid combinations of character profiles, relationships, etc., take place for all episodes under the same scenario. For this study, we focused only on two Sotopia scenarios selected to explore the impacts of personality trait and AI capability settings on negotiation outcomes. Experiment 1 focuses on a price bargaining negotiation between two agents. Experiment 2 focuses on a job negotiation scenario between an AI hiring manager and a simulated human job candidate.

\subsubsection{Character profiles} Sotopia character profiles are defined using key traits that influence agents' behavior and decision-making during social interactions. These include public attributes such as name, gender, occupation, relationships, and moral values, as well as private information akin to real-life secrets. Two character profile examples are shown in Figure~\ref{fig:sotopia-framework}. In this study, we initialized our agents using default Sotopia character templates from \cite{zhouSOTOPIAInteractiveEvaluation2024} and made targeted modifications to their Big Five personality traits to align with the specific goals of each of our experiments.  Additionally, although Sotopia allows for scenarios to take place with five possible relationship types between characters, this study only considers negotiation simulations between strangers. We selected this constraint given the nature of the selected scenarios generally taking place between strangers in real-life settings. 

\subsubsection{Social goals} Sotopia social goals serve as the primary motivating factors behind each agent’s behaviors throughout an episode. These goals are private to each agent, akin to characters' individual secrets, and may or may not be in conflict with their interaction partner's respective social goals. While simulated AI agents do not have character profiles, they also have social goals.

\subsubsection{AI characteristics} In Sotopia scenarios where one of the agents is a simulated AI agent instead of a human, such as Experiment 2 of this study, character profiles are replaced by AI characteristics. Unlike character profiles, these are direct manipulations of simulated AI agents' communication capabilities, namely their transparency, competence, and adaptability. 

\subsection{Evaluation Measures}
We used three measure categories to explore how personality traits and AI characteristics impact our Sotopia negotiation simulations.

\begin{table}[!t]
    \small
    \centering
    \begin{tabular}{@{}p{5.5cm} p{10cm}@{}}
        \toprule
        \textbf{Sotopia-Eval Dimension} & \textbf{Description [Score Range]} \\
        \midrule
        Goal Completion & How well the agent achieves its defined social goals \textnormal{[0, 10]}. \\
        Believability & How natural, realistic, and consistent the agent’s behavior is with its character profile \textnormal{[0, 10]}. \\
        Knowledge Acquisition & Agent’s success in acquiring new, relevant, and important information \textnormal{[0, 10]}. \\
        Secret Keeping & Extent to which agent maintains secrecy of private information or intentions \textnormal{[$-$10, 0]}. \\
        Relationship Change & How the interaction influences the agent’s relationships and social reputation \textnormal{[$-$5, 5]}. \\
        Social Rule Compliance & Adherence to legal rules and social norms during the interaction \textnormal{[$-$10, 0]}. \\
        Financial and Material Benefits & Gains in monetary or material terms, both short- and long-term \textnormal{[$-$5, 5]}. \\
        \bottomrule
    \end{tabular}
    \caption{Sotopia-Eval agent interaction dimensions~\cite{zhouSOTOPIAInteractiveEvaluation2024}}
    \label{tab:sotopia_dimensions}
\end{table}

\begin{table}[t!]
\small
    \centering
    \begin{tabular}{p{0.2\linewidth} p{0.7\linewidth} p{0.1\linewidth}}
    \toprule
         \textbf{Category} & \textbf{Constructs} &  \\ 
    \midrule
         \multirow{2}{*}{Empathy} & Empathy Intent & ~\cite{see2019makes} \\
          & Empathy Emotions & ~\cite{see2019makes} \\
    \midrule
         \multirow{3}{*}{Socio-cognitive} & Connotations, Perspectives, Attitudes & ~\cite{rashkinConnotationFramesDataDriven2016} \\
          & Moral Values (Harm, Fairness, Purity, Authority, Ingroup) & ~\cite{
          graham2013moral} \\
          & Subjectivity & ~\cite{rashkinTruthVaryingShades2017} \\
    \midrule
         \multirow{3}{*}{Emotional} & Sentiment & ~\cite{sanhDistilBERTDistilledVersion2020} \\
          & Toxicity & ~\cite{Detoxify} \\
          & Emotions & ~\cite{savaniDistilBERTEmotionRecognition2024} \\
    \bottomrule
    \end{tabular}
    \caption{Lexical measures from communications~\cite{volkova2021machine}}
    \label{tab:SEC}
\end{table}

\subsubsection{Scenario-based measures} We derived scenario-based measures from Sotopia-Eval, a multidimensional evaluation scale developed specifically according to Sotopia interpersonal social simulation parameters (Table~\ref{tab:sotopia_dimensions}; \citep{zhouSOTOPIAInteractiveEvaluation2024}). These captured intervention impacts relative to simulation outcomes, such as the completion of social, material, or knowledge goals, as well as the qualities of the interactions comprising a simulation episode.

\subsubsection{Lexical measures} We used a suite of AI-driven and lexicon-based cognitive domain analytics (Table~\ref{tab:SEC}) to capture the extent to which Sotopia episodes approximated linguistic markers of social, cognitive, and emotional processes in our social simulations. These included sentiment \cite{savaniDistilBERTEmotionRecognition2024}, toxicity \cite{hanuDetoxify2020}, empathy with others' emotions and intents \cite{leeDoesGPT3Generate2022}, emotions \cite{devlinBERTPretrainingDeep2019}, moral values \cite{garten2016morality}, connotation frame analysis \cite{rashkinConnotationFramesDataDriven2016a}, subjectivity \cite{rashkinTruthVaryingShades2017}, and hate \cite{aluruDeepDiveMultilingual2021}.

\subsubsection{Questionnaire measures} Finally, we administered questionnaires at the end of episodes to serve as analogs of surveys for measuring reflective measures of various socio-cognitive constructs in human subjects experiments~\cite{zhouSOTOPIAInteractiveEvaluation2024}. 

\section{Experiment 1: Interpersonal Price Negotiation}
\subsection{Method}


        

\subsubsection{Scenario}
For this experiment, our simulation scenario is a bargaining task based on detailed product description and target prices from fictitious Craigslist deals \cite{he2018decoupling}. The objective of the agents is to strike a deal while getting as close to their own target price as possible. In total, we include 10 scenarios featuring different items for negotiation.

\subsubsection{Measures}
For this experiment, we measured only scenario-based and lexical measures. Scenario-based measures used the seven original dimensions of the Sotopia-Eval evaluation scale (Table~\ref{tab:sotopia_dimensions}), namely: Believability, Financial and Material Benefits, Goal, Knowledge, Overall Score, Relationship, Secret, and Social Rules. We also employed our lexical measures to describe the conversational dynamics of simulated socializations, particularly indicators of agents' empathy with respect to an interaction partner's emotions and intents; moral foundation indicators; positive and negative sentiments, including the use of emotional vocabulary; and the use of connotation frames---subtle lexical markers indicating implied perspectives, presupposed values, resulting effects, and likely mental state associated with entities involved in an event \citep{rashkinConnotationFramesDataDriven2016a}.

\subsubsection{Experiment Settings}
We used gpt-4o-mini for all agents, running a total of 4343 episodes for each treatment combination setting to ensure stability, generating a total of 8686 transcripts. The temperatures are set to 0.7 to ensure consistency.

\subsubsection{Causal Investigations of Simulations}
We employ two techniques to analyze the impacts of scenario type and personality interventions in this experiment. First, we leveraged causal discovery approaches to: (1) explore causal linkages and structure (i.e., ``when \textit{X} increases, we see a decrease in \textit{Y}''); and (2) estimate average treatment effects (ATEs) on a target metric as a result of an intervention. This approach follows Pearl and Mackenzie's \cite{pearlBookWhyNew2018} causal analytic framework. We specifically used CausalNex \cite{beaumontCausalNex2021} to create directed acyclic graphs (DAGs) from Sotopia simulation outputs, in which intervention and outcome variables are represented by node and edge relationships with no fully closed loops. Following the causal structure learning step, we estimated the ATEs, which represent the average differences between treated and non-treated samples. To do so, we used Causal Forests \cite{atheyGeneralizedRandomForests2019b} as applied in the Python project EconML \cite{battocchiEconMLPythonPackage2019} to identify interventions and outcome of choice then estimate the ATE, repeating for all intervention-outcome pairs, while removing other outcomes for bias control. 


\subsection{Results}



 

\subsubsection{Scenario-based Measures}
\begin{figure}[t]
    \centering
    \includegraphics[width=0.49\linewidth]{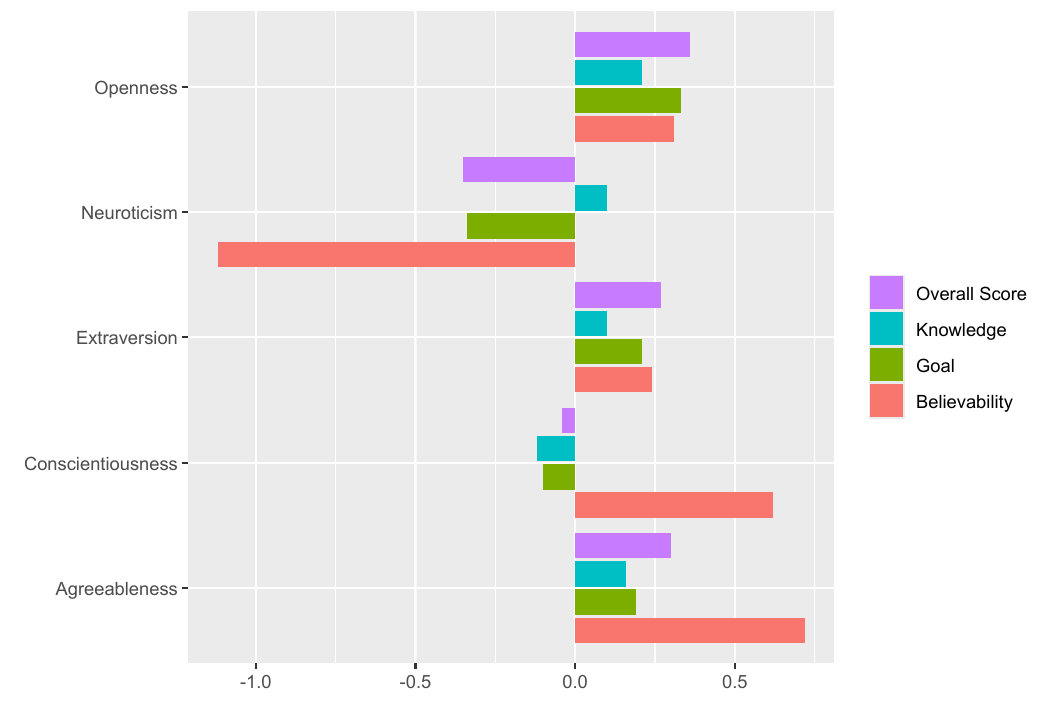}
    \caption{Trait level--Sotopia-Eval SEM Weights}
    \label{fig:sotopia2-sotopia}
\end{figure}


Sotopia-Eval measure findings (Figure~\ref{fig:sotopia2-sotopia}) showed a positive association between personality trait levels and all measures, with the opposite trend observed for neuroticism. Personality trait treatments only significantly impacted Believability, Goal, Knowledge, and Overall Score. Among these, Believability was the most consistently impacted by personality trait level manipulations, and Knowledge was the least. 

\subsubsection{Lexical Measures: Empathy}

\begin{figure}[t]
\centering

\begin{subfigure}[t]{0.49\textwidth}
    \includegraphics[width=\linewidth]{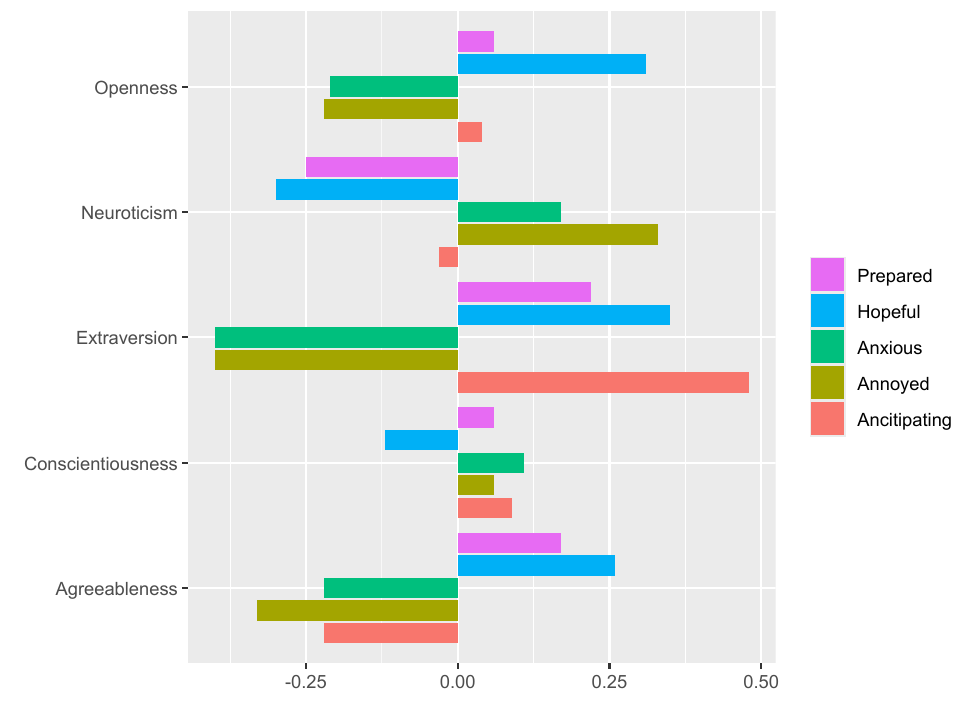}
    \caption{Empathy Emotion Measures}
    \label{fig:sotopia2-empathy-emotion}
\end{subfigure}
\hfill
\begin{subfigure}[t]{0.49\textwidth}
    \includegraphics[width=\linewidth]{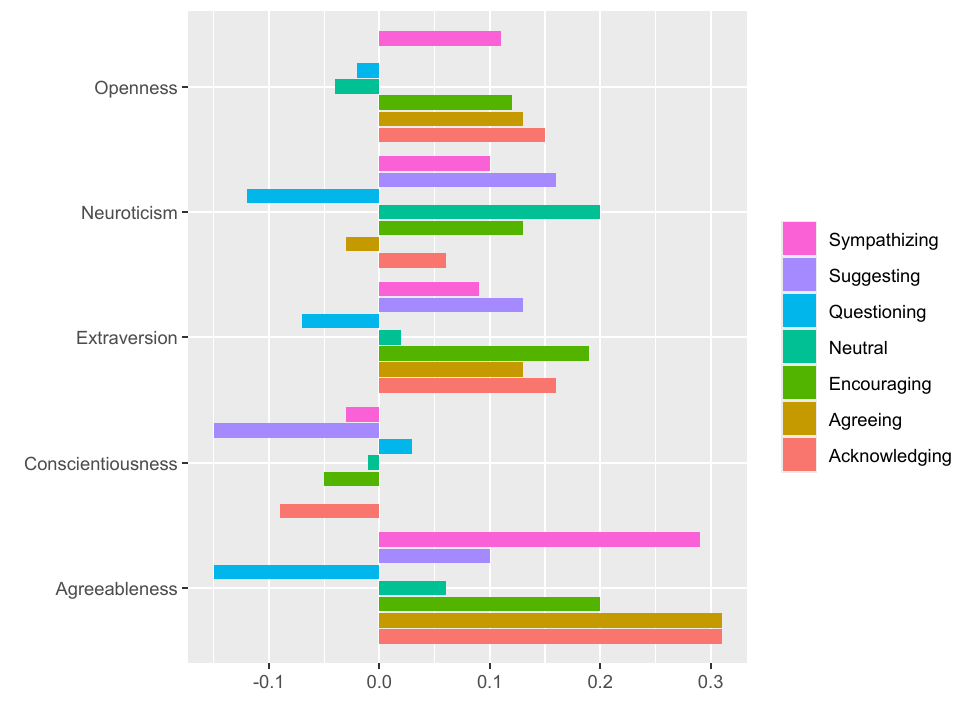}
    \caption{Empathy Intent Measures}
    \label{fig:sotopia2-empathy-intent}
\end{subfigure}

\caption{Trait level--Empathy lexical measure SEM Weights}
\label{fig:sem-empathy-subfigures}
\end{figure}

Personality trait treatments resulted in appreciable SEM weight impacts across several lexical measures, particularly empathic speech markers. For emotion-specific empathy measures (Figure~\ref{fig:sotopia2-empathy-emotion}), only five emotional empathy markers were significantly affected by personality treatments: Annoyed, Anticipating, Anxious, Hopeful, and Prepared. We found the largest effects on Hopeful, Anxious, and Annoyed emotional empathy markers, with Annoyed and Anxious sharing similar and opposite patterns as Hopeful and Prepared. Extraversion treatment levels produced the largest effects across all emotional empathy measures.

Significant impacts were also found on seven intent-based empathy indicators: Sympathizing, Suggesting, Questioning, Neutral, Encouraging, Agreeing, and Acknowledging (Figure~\ref{fig:sotopia2-empathy-intent}). Intent empathy marker impacts were generally positively weighted across personality trait level manipulations, with the exception of Conscientiousness. However, this trend was notably reversed for markers of empathy with Questioning intents.

\subsubsection{Lexical Measures: Moral Foundations}
\begin{figure}
    \centering
    \includegraphics[width=0.49\linewidth]{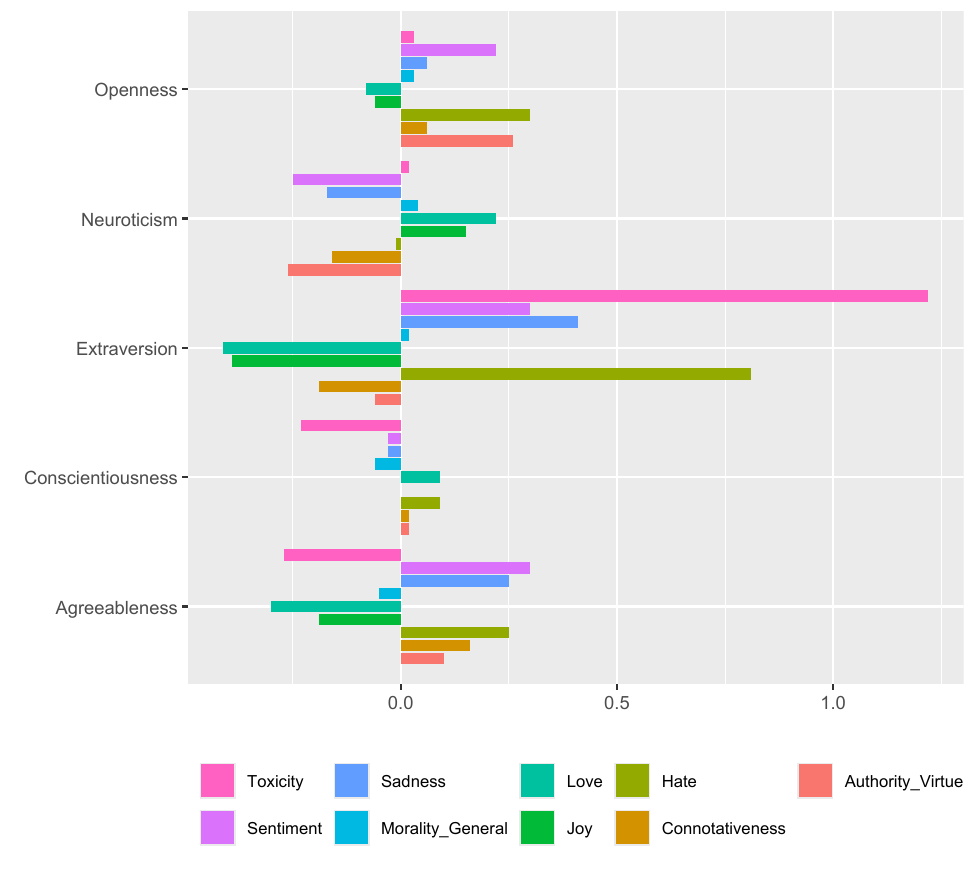}
    \caption{Trait level--Socio-cognitive-emotion SEM Weights}
    \label{fig:sotopia2-other-measures}
\end{figure}

Of the five moral foundation measures, only Morality\_General and Authority\_Virtue-related measures were appreciably impacted by personality trait treatments (Figure~\ref{fig:sotopia2-other-measures}). 
Authority\_Virtue, which measures affirmative references to hierarchical social structures, was positively associated with Agreeableness, Conscientiousness, and Openness levels, and negatively associated with the other two traits. 

\subsubsection{Lexical Measures: Sentiment, Emotion, and Toxicity}
Lexical indicators of emotionally-charged language were significantly impacted by our personality treatments (Figure~\ref{fig:sotopia2-other-measures}). Overall Sentiment scores were positively correlated with Openness, Extraversion, and Agreeableness levels, and negatively associated with the other two traits. The same trends were found for Hate and Sadness indicators, which were reversed for for Love and Joy indicators. Toxicity scores, which consider potentially humorous or sarcastic uses of hateful vocabulary, followed similar personality trait level correlation trends as Hate and Sadness scores, with the exception of Agreeableness. Extraversion produced the most extreme SEM weight impacts across these five measures.

\subsubsection{Lexical Measures: Connotation}
Finally, we also found significant impacts of personality trait treatments on the general use of connotation frames. Figure~\ref{fig:sotopia2-other-measures} shows that the strongest effects we found were a positive personality trait level correlation difference with Agreeableness and a negative one for Extraversion. Smaller, positive trait level differences were found for Conscientiousness and Conscientiousness and Openness. Despite a marginal trait level difference, controlling for Neuroticism appears to generally reduce the use of connotation frames.

\subsection{Discussion}

Experiment 1 findings generally aligned with established literature on Big Five personality trait theories, especially regarding social and negotiation contexts. Scenario-based measures supported the positive roles of Agreeableness, Extraversion, and Openness, consistent with known relationships linking Agreeableness to cooperation, trust, and prosocial behaviors \citep{bradleyTeamPlayersCollective2013, driskellWhatMakesGood2006}, Extraversion to assertive communication and social engagement \citep{barry1998bargainer, grantRethinkingExtravertedSales2013}, and Openness to adaptability and openness toward new information during collaborative tasks \citep{lepineAdaptabilityChangingTask2000, bellDeeplevelCompositionVariables2007}. Conversely, the negative effects of Neuroticism on interaction outcomes matched prior findings associating neurotic traits with emotional instability and interpersonal conflict \citep{driskellWhatMakesGood2006, kleinHOWTHEYGET2004}. The limited role of Conscientiousness, primarily affecting believability, also aligns with literature suggesting that conscientious behaviors are less salient in brief conversational interactions, requiring further validation in LLM-based social simulations \citep{peetersPersonalityTeamPerformance2006}.

Lexical analyses reinforced personality theory expectations regarding emotional responsiveness, moral expression, and nuanced communication strategies in negotiation contexts. Empathy markers demonstrated clear trait-linked patterns consistent with known emotional expressivity and optimism associated with Extraversion \citep{watson1994panas, costa2008revised, mccraeNEOPIRData362002}, and prosocial empathic responses linked to Agreeableness and Openness \citep{grazianoAgreeablenessEmpathyHelping2007, habashiSearchingProsocialPersonality2016}. Findings interpreted via Moral Foundations theory further indicated personality-driven differences in moral communication, with Conscientiousness strongly linked to structured, rule-oriented interactions, and Neuroticism associated negatively with authority-related moral expressions \citep{grahamLiberalsConservativesRely2009, hirshCompassionateLiberalsPolite2010}. Lastly, sentiment and connotation framing analyses confirmed known associations of Agreeableness and Extraversion with positive affectivity, interpersonal warmth, and direct or subtle communicative styles, respectively \citep{mccraeIntroductionFiveFactorModel1992, watsonExtraversionItsPositive1997, rashkinConnotationFramesDataDriven2016a, pennebakerLinguisticStylesLanguage1999}. Together, these lexical results substantiate the validity of simulated personality manipulations in capturing established human social phenomena within LLM-driven negotiation scenarios.

While Experiment 1 validated the capability of LLM-driven simulations to reflect established personality trait effects in human negotiation scenarios, Experiment 2 aims to extend these findings into negotiation contexts involving AI agents. Thus, Experiment 2 investigates how AI Transparency, Adaptability, and Reliability interact with key personality traits identified previously---Agreeableness and Extraversion---in shaping social negotiations.
\section{Experiment 2: Human-AI Job Negotiation}
\label{sec:exp2}


\subsection{Method}

\subsubsection{Scenario} 
Our second experiment aimed to understand how AI Agent traits influence negotiation outcomes alongside the most influential Big Five personality traits from Experiment 1, namely Agreeableness and Extraversion. We employed a scenario in which an AI Bot hiring manager negotiates with a human digital twin (HDT) job candidate over key terms of a job offer, such as the start date and salary. Each key negotiation term has five evenly spaced options (e.g., salary options from \$100k to \$120k in increments of \$5k), with each option corresponding to a fixed number of points for the AI hiring manager and the simulated human job candidate. The point designations are inversely proportional, creating a zero-sum dynamic where one agent’s gain directly reduces the other’s score. For example, if the final salary is \$120k, the candidate receives 6000 points, while the recruiter receives 0 points; a lower final salary would increase the recruiter's points at the expense of the candidate's. The detailed scoring table is shown in Appendix \ref{appendix:job_negotiation}.

In addition to varying the personality traits of the HDT job candidates, we investigated how AI Bot characteristics influence negotiation measures by manipulating the hiring manager's interaction traits along three dimensions: Transparency, Adaptability, and Reliability---each with High and Low variations. The exact prompt formulations for each trait variation are provided in Appendix \ref{appendix:trait_variation}.

\subsubsection{Experiment Settings}
We used gpt-4o \footnote{\url{https://openai.com/index/hello-gpt-4o}} for both AI bot and job candidates, running 20 episodes for each treatment combination setting to ensure stability. The temperatures are set to 0.7 to ensure consistency. We generated 1,280 job negotiation transcripts.

\subsubsection{Measures} 
As with Experiment 1, we collected scenario- and lexical-based measures to systematically analyze intervention impacts within the simulated negotiation scenario. 

We focused on four main Scenario-based measures for this experiment. \textit{Deal-making success} was a binary indicator for whether negotiation concluded with an agreement. \textit{Negotiation points} were distributed between the recruiter and candidate, assigned to each following a zero-sum framework to create realistic trade-offs in the negotiation. \textit{Transactivity} measured the frequency of transactive exchanges between agents, weighted according to the elaboration, idea building, questioning, and argumentation involved. \textit{Verbal Equity} measured the extent to which agent interactions showed a balanced distribution of speaking opportunities among agents.

As in Experiment 1, we also included lexical-based measures, such as markers of empathy with others' intent and emotion, moral values, sentiment, and emotions. We also analyzed simulated negotiations for lexical indicators of socio-cognitive states, particularly in the form of connotative language use and subjectivity word usage. 
Finally, we administered questionnaire-based measures after each simulation to simulate participant responses to post-experimental surveys in human subject research databases. 

\subsubsection{Causal Investigations of Simulations}
The same causal discovery approaches used in Experiment 1 was used to investigate the impacts of personality interventions on all outcome variables for this experiment. 

\subsection{Results}

\subsubsection{Scenario-based Measures}

\begin{figure}
    \centering
    \includegraphics[width=0.6\linewidth]{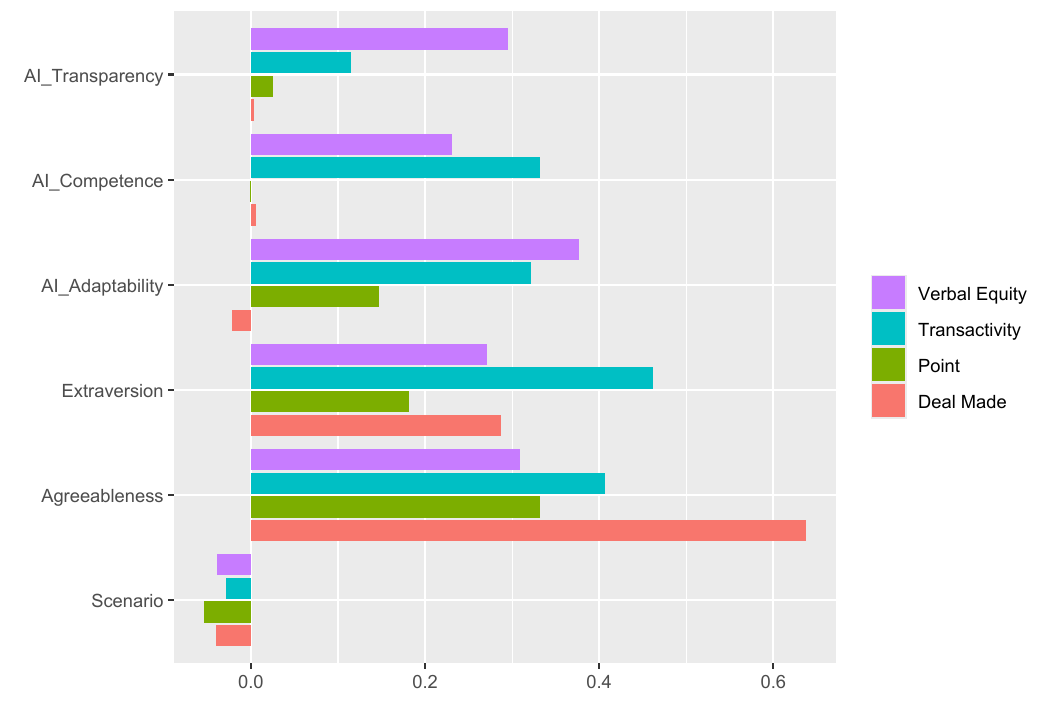}
    \caption{Scenario-based measure SEM Weights}
    \label{fig:job-scenario-measures}
\end{figure}

Our results suggest that both HDT personality traits and AI bot characteristics play a crucial role in shaping several qualities of simulated job negotiation interactions (Figure~\ref{fig:job-scenario-measures}). AI transparency, competence, and adaptability produced moderately strong positive associations with transactivity, indicating that interactions become more dynamic, engaging, and reciprocal. Similar positive associations were found for verbal equity, which reflects a balanced and fair exchange of dialogue. 

\subsubsection{Questionnaire measures}

\begin{figure}
    \centering
    \includegraphics[width=0.6\linewidth]{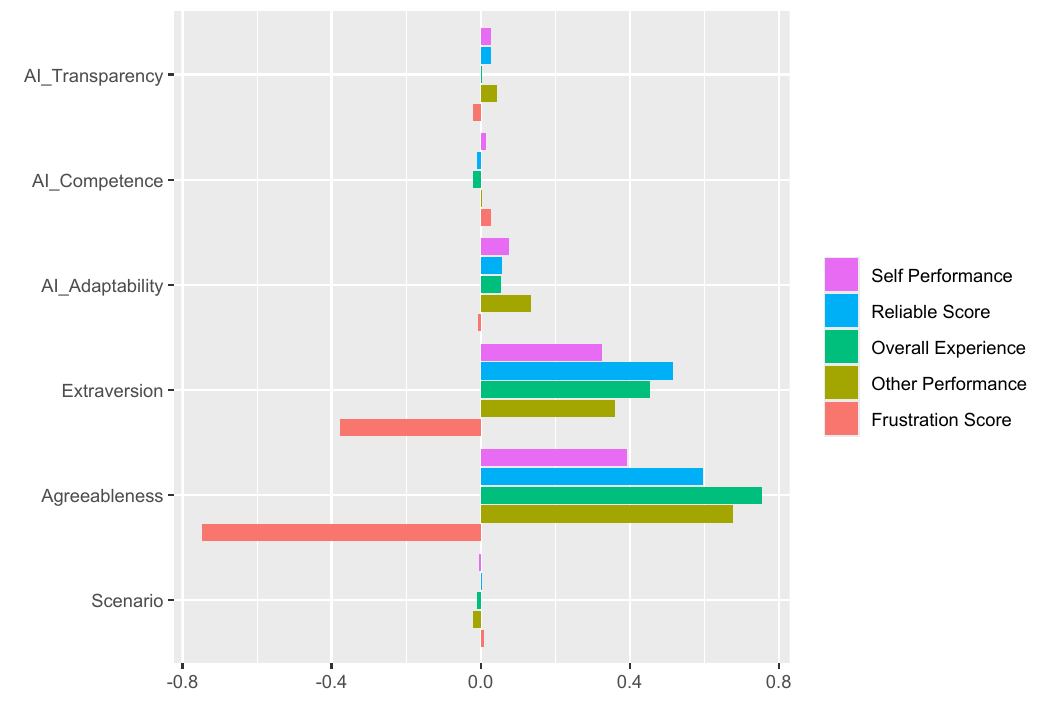}
    \caption{Questionnaire measure SEM Weights}
    \label{fig:job-survey-measures}
\end{figure}

We found notable significant impacts of HDT personality and AI  traits on questionnaires meant to resemble post-experimental surveys to evaluate participants' experiences interacting with an AI agent (Figure~\ref{fig:job-survey-measures}). Agreeableness and Extraversion were strongly and positively associated with several questionnaire measures, with the exception of frustration scores.
In contrast, AI Bot only marginally influenced HDTs' questionnaire responses, with AI adaptability having the strongest (and most positive) impacts. 

\subsubsection{Lexical measures: Empathy}

\begin{figure}
    \centering
    \includegraphics[width=0.6\linewidth]{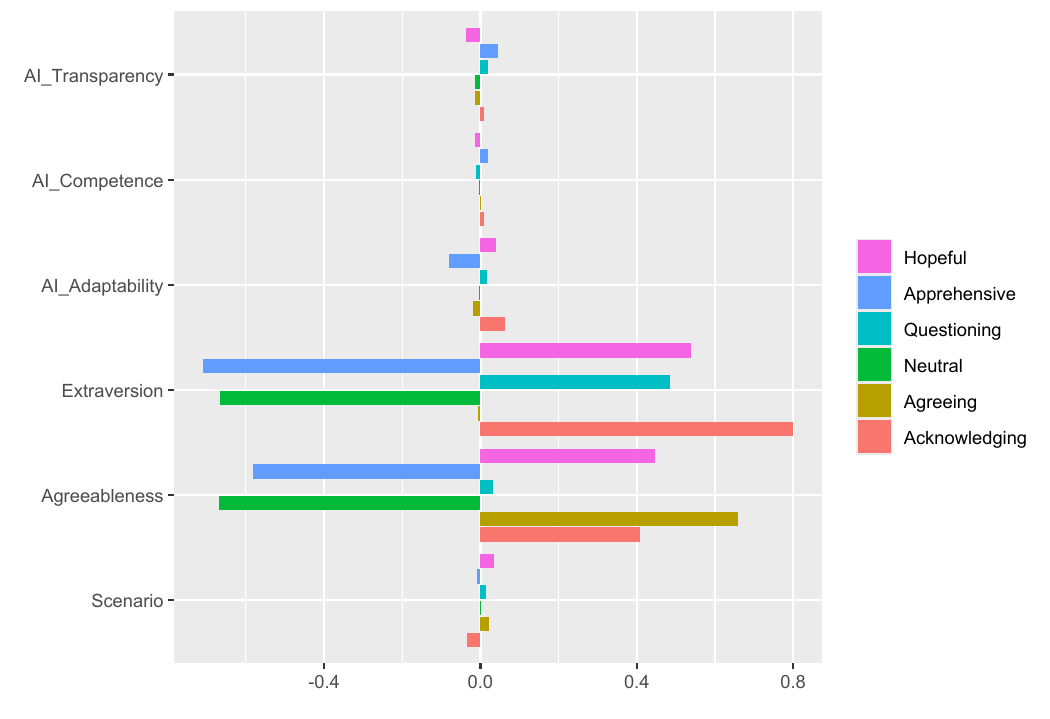}
    \caption{Empathy measure SEM Weights: ``Hopeful'' and ``Apprehensive'' are emotion empathy measures, and the rest are intent empathy measures}
    \label{fig:job-empathy-measures}
\end{figure}

We found an outsize impact of HDT personality trait levels compared to AI characteristics across all empathy markers (Figure~\ref{fig:job-empathy-measures}). Emotional empathy findings were consistent with Experiment 1: lexical markers for empathizing with hopeful emotions was positively associated with both Extraversion and Agreeableness, while negative trends were found for apprehensiveness empathy markers. With the exception of AI Adaptability's impact on the Apprehensive measure, AI Bot transparency, competence, and adaptability levels did not appreciably impact emotional empathy measures.

Intent empathy measures were less consistent with our Experiment 1 findings, with only the Acknowledging intent measure exhibiting positive associations with Extraversion and Agreeableness. Some of our intent empathy findings were as expected: Extraversion was positively associated with markers of empathizing with others' Questioning beliefs, while Agreeableness was positively associated with our Agreeing empathy measure. In contrast to our Experiment 1 findings, both personality treatments had moderately strong negative effects on the Neutral empathy measure. AI Bot treatments did not significantly impact any intent empathy measure.

\subsubsection{Lexical measures: Morality, Sentiment, and Emotion}

\begin{figure}
    \centering
    \includegraphics[width=0.6\linewidth]{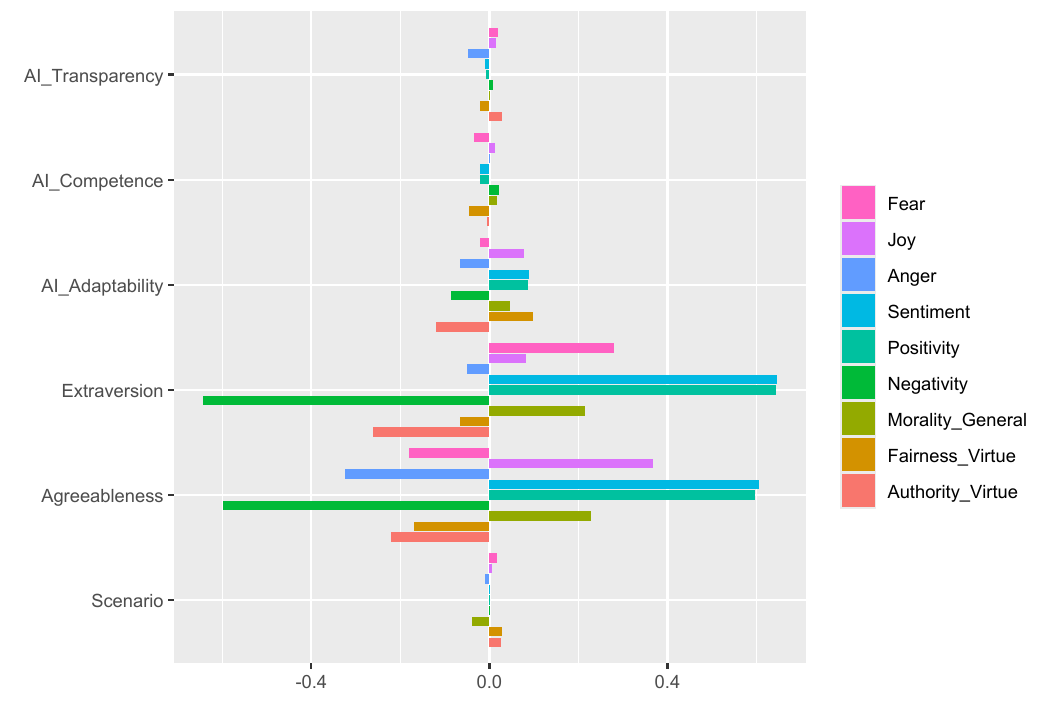}
    \caption{Moral Foundation, Sentiment, and Emotion measure SEM Weights}
    \label{fig:job-empathy-measures}
\end{figure}

HDT Extraversion and Agreeableness produced much stronger effects on lexical measures of interactions' inclusion of moral foundation, sentiment, and emotion markers, compared to AI Bot treatment levels (Figure~\ref{fig:job-empathy-measures}). As with Experiment 1, both personality treatments were significantly positively associated with Joy, Overall Sentiment, and Positivity---as well as Morality\_General. In contrast, lexical measures of Anger, Negativity, and positive views on Fairness and Authority (``Fairness\_Virtue'' and ``Authority\_Virtue'') were negatively associated with higher Extraversion and Agreeableness levels. Fear, a new emotion measure we used for Experiment 2, was positively associated with Extraversion but negatively associated with Agreeableness. Among AI Bot trait level treatments, only AI Adaptability had a significant effect on any of these measures: a weak negative association on Authority\_Virtue.

\subsubsection{Lexical measures: Connotative Framing}





\begin{figure}[!t]
    \centering
    \includegraphics[width=0.6\textwidth]{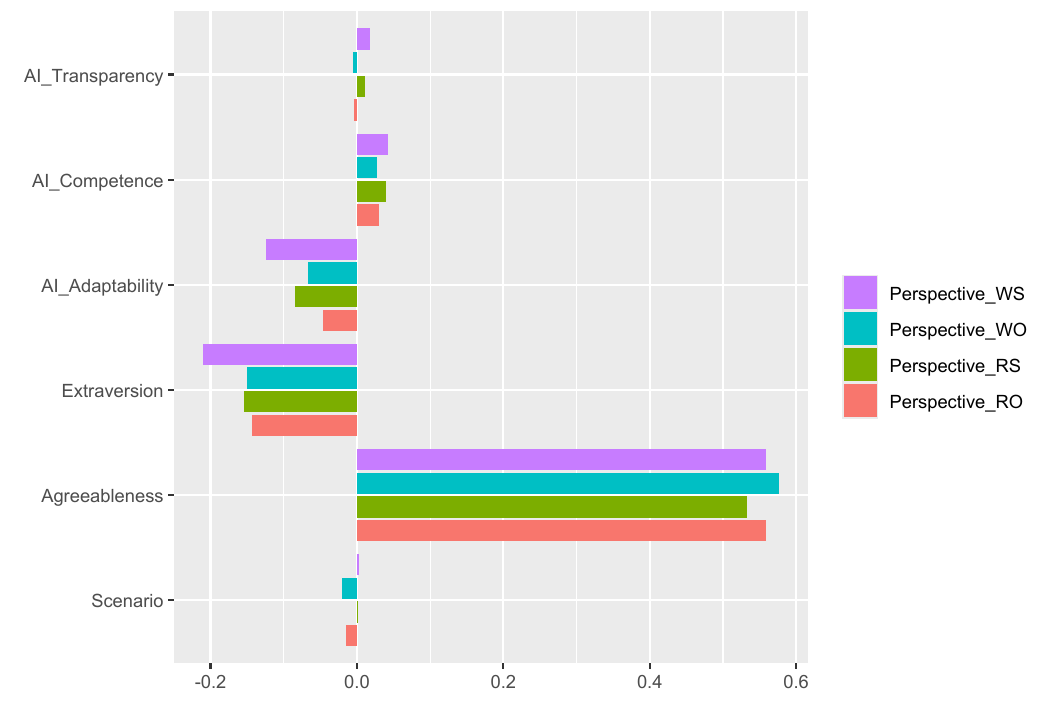}
    \caption{Lexical Connotation Frame Measures. Suffix indicates whose perspective is being implied (\textit{\textbf{W}}riter/\textbf{\textit{R}}eader) and the topic of implied sentiment (\textbf{\textit{S}}ubject/\textbf{\textit{O}}bject).}
    \label{fig:jobNegotiation-connotation}
\end{figure}

Connotative framing findings (Figure~\ref{fig:jobNegotiation-connotation}) indicate that only HDTs' Agreeableness levels had moderately strong impacts on all forms of connotative markers: high Agreeableness HDTs used more connotative language across the board. Significant but weak negative effects of Extraversion levels were also found on all forms of connotativeness. Similar, but notably weaker, negative trends were found for AI Adaptability---the only AI Bot treatment to appreciably impact the usage of connotative language during the negotiation episodes.


\subsubsection{Lexical measures: Subjectivity}

\begin{figure}[!t]
    \centering
    \includegraphics[width=0.6\textwidth]{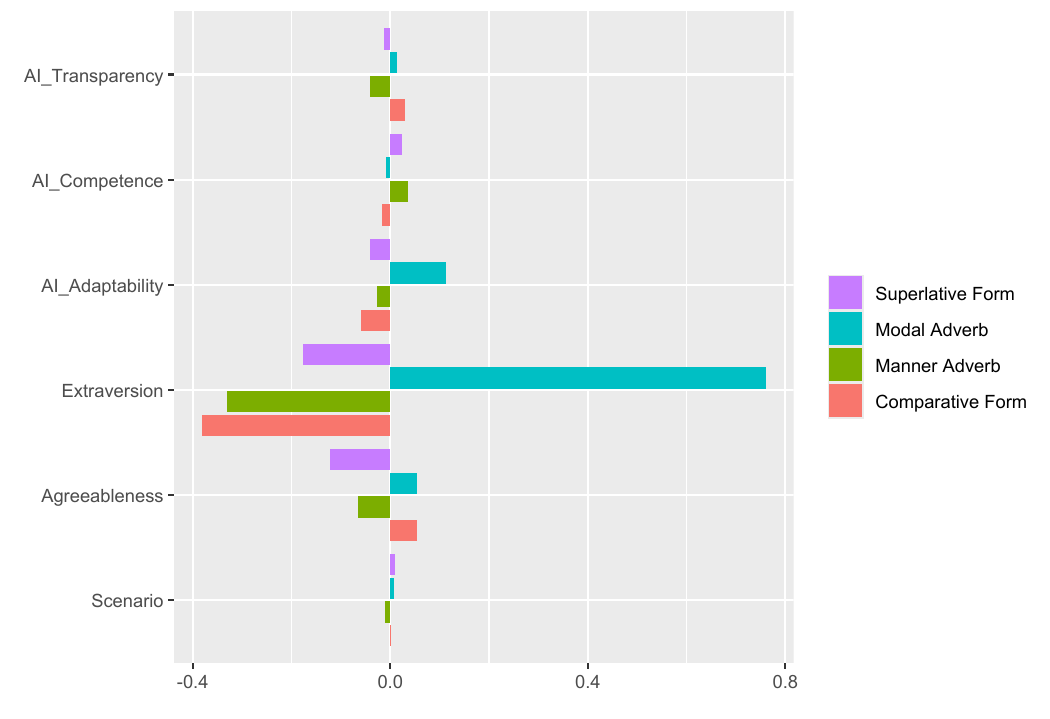}
    \caption{Subjectivity measure SEM weights}
    \label{fig:jobNegotiation-subjectivity}
\end{figure}

Measures of subjective or evocative language (Figure~\ref{fig:jobNegotiation-subjectivity}) were only consistently impacted by HDT Extraversion levels. We found that high Extraversion and Agreeableness had significant but weak negative impacts on the usage of superlative adverbs (e.g., ``\textbf{most} valuable''). Slightly stronger negative impacts of high Extraversion were found on the use of manner adverbs (i.e., indicators of \textit{how} an action is done, such as ``\textbf{happily}'' or ``\textbf{slowly}'') and comparative phrases (e.g., ``\textbf{more} \_\_\_ \textbf{than}''). Extraversion had a markedly strong positive effect on Modal adverb usage (i.e., indicators of uncertainty, such as ``\textbf{probably}''). AI Adaptability also had a weak positive effect on Modal adverb use, which was the only appreciable AI Bot treatment effect on subjectivity lexical measures.

\subsection{Discussion}
Experiment 2 expands on our previous findings on LLM-based social negotiation simulations by examining AI Bot Transparency, Competence, and Adaptability alongside simulated human negotiators' Agreeableness and Extraversion levels. Scenario-based results showed that higher AI Transparency, Competence, and Adaptability positively impacted transactivity and verbal equity. These results align with existing research emphasizing the importance of transparency and adaptability in human-AI interactions, highlighting that clearer but adaptive communication capabilities facilitate more reciprocal and balanced interactions \citep{hancockEvolvingTrustRobots2021, schaeferMetaAnalysisFactorsInfluencing2016}. The moderate but consistent impacts of AI characteristics underline the role of AI agent qualities in shaping conversational dynamics, especially in negotiation settings where balanced participation is essential \citep{barry1998bargainer, curhanWhatPeopleValue2006}.

Questionnaire measures reinforced personality-driven findings from Experiment 1. The strong positive association of Agreeableness and Extraversion with questionnaire measures regarding the AI Bot's reliability, honesty, and effort corresponds closely to prior research linking Agreeableness and Extraversion with cooperative engagement, trustworthiness, and positive affective experiences in interpersonal negotiation contexts \citep{grazianoAgreeablenessEmpathyHelping2007, grantRethinkingExtravertedSales2013, dimotakisMindHeartLiterally2012}, as well as general human-agent interactions \cite{hancockMetaAnalysisFactorsAffecting2011, schaeferMetaAnalysisFactorsInfluencing2016, jessupMeasurementPropensityTrust2019}. In contrast, AI characteristics influenced questionnaire outcomes only marginally, with AI Adaptability showing the most notable, though still limited, effect. This aligns with existing literature suggesting that participants' subjective experiences in human–AI interactions are more sensitive to perceived interpersonal qualities (e.g., warmth or openness) than technical features alone \citep{prajodEffectVirtualAgent2019}.

Lexical analyses further validated personality trait effects observed in Experiment 1. Empathy findings remained largely consistent with Experiment 1 and established literature, showing positive associations of Extraversion and Agreeableness with emotional expressiveness and supportive, prosocial empathic responses \citep{watson1994panas, grazianoAgreeablenessEmpathyHelping2007}. Interestingly, intent-based empathy markers differed slightly from Experiment 1, with unexpected negative associations for neutral empathic intents, potentially reflecting situational nuances in negotiation interactions. Connotation framing analyses supported earlier evidence linking higher Agreeableness to increased use of subtle, polite, or nuanced communication styles, while higher Extraversion was associated negatively due to extraverts' tendency toward more direct and less implicitly nuanced communication \citep{rashkinConnotationFramesDataDriven2016a, pennebakerLinguisticStylesLanguage1999}. These findings collectively underscore the robustness of personality effects on negotiation behaviors across LLM simulations, while potentially indicating a gap in simulating the complexity of human-AI socializations.

\section{General Discussion and Conclusion}

Across two experiments, we demonstrated the effectiveness of large language model (LLM)-driven simulations in modeling personality-driven dynamics within negotiation scenarios. Experiment 1 provided strong evidence that personality trait prompts for Agreeableness, Extraversion, Openness, and Neuroticism produce simulated behaviors consistent with established theoretical predictions and empirical findings in the personality and negotiation literature \citep{barry1998bargainer, bradleyTeamPlayersCollective2013, grazianoAgreeablenessEmpathyHelping2007, driskellWhatMakesGood2006}. Lexical analyses revealed that personality traits systematically influenced emotional expressivity, moral expressions, and nuanced lexical patterns in ways aligning closely with previously observed human negotiation behaviors \citep{pennebakerLinguisticStylesLanguage1999, rashkinConnotationFramesDataDriven2016a, grahamLiberalsConservativesRely2009}. Thus, Experiment 1 underscores the utility of LLMs for controlled, scalable investigation into personality-driven interpersonal dynamics.

Experiment 2 built on these findings by exploring the joint impacts of human digital twin (HDT) personality traits and AI agent characteristics—Transparency, Competence, and Adaptability—on simulated job negotiation outcomes. The results suggested that AI agent characteristics, particularly Adaptability and Transparency, influenced interaction dynamics such as transactivity and verbal equity, although these impacts were moderate compared to those driven by HDT personality traits. Questionnaire outcomes and lexical analyses consistently showed strong effects of Agreeableness and Extraversion on participants' subjective negotiation experiences and conversational behaviors, echoing Experiment 1 results and further highlighting that interpersonal traits substantially shape negotiation interactions and outcomes \citep{dimotakisMindHeartLiterally2012, watson1994panas, grazianoAgreeablenessEmpathyHelping2007}. In contrast, AI agent traits played a largely marginal role influencing conversational balance and subtle interaction nuances, aligning with literature suggesting that human-AI interactions depend significantly on perceived interpersonal and social qualities \citep{hancockEvolvingTrustRobots2021, schaeferMetaAnalysisFactorsInfluencing2016, prajodEffectVirtualAgent2019}.

Taken together, our findings provide evidence for the feasibility of employing LLM-based social simulations as valid platforms for investigating complex personality-driven dynamics in negotiation and HAT contexts. The observed alignment of simulated behaviors with existing empirical evidence suggests the promise of LLMs for systematically exploring nuanced interpersonal and communicative phenomena. Our findings also point toward directions for future work: examining interactions between personality traits at finer granularity, systematically exploring additional AI-agent traits, and extending analyses across other types of social interaction tasks. Ultimately, this approach offers researchers a highly scalable and controllable experimental framework for refining theories and practical strategies around personality-informed design in human-AI interactions.
\section{Limitations}
Despite the novel insights provided by this work, several limitations should be acknowledged. First, while our simulations demonstrated alignment with established personality theories, they rely on prompt-based personality manipulations that may not fully capture the complexity of human personality expression in real-world negotiations. Second, our experiments focused on specific negotiation scenarios (price bargaining and job negotiations), which may limit generalizability to other mission-critical contexts such as crisis management or tactical coordination. Third, the lexical measures, while comprehensive, depend on the quality of LLM-generated dialogue and may not capture non-verbal cues critical to human negotiation dynamics. Finally, our AI agent characteristics were limited to transparency, competence, and adaptability, potentially overlooking other crucial factors that influence human-AI teaming effectiveness in operational environments.

\section{Operational Implications}
Our findings have direct implications for deploying agentic AI systems in defense and mission-critical operations. The strong causal effects of Agreeableness and Extraversion on negotiation outcomes suggest that AI agents must be designed to recognize and adapt to operator personality profiles in real-time. For defense applications, this means developing AI systems capable of adjusting their communication strategies when interfacing with diverse military personnel, coalition partners, or civilian stakeholders. The dominance of personality effects over AI characteristics indicates that training protocols should emphasize personality-aware interaction design rather than purely technical enhancements. Furthermore, our multi-dimensional evaluation framework provides a blueprint for pre-deployment testing of AI agents, enabling commanders to assess whether specific AI systems will perform effectively with their particular team compositions. 

\section{Future Work}
Building on these foundational findings, we plan to extend our research in several critical directions. First, we will conduct additional experiments examining competitive versus collaborative job negotiation scenarios to understand how task framing influences the interaction between personality traits and AI characteristics. This distinction is particularly relevant for defense applications where AI agents must seamlessly transition between cooperative allied interactions and competitive adversarial negotiations. Second, we will expand our AI characteristic framework to include warmth and theory of mind capabilities, as these factors are essential for building trust and mutual understanding in high-stakes human-AI teams. Warmth, in particular, may moderate the effects of personality traits on negotiation outcomes and could be crucial for AI agents operating in culturally diverse environments. Third, we plan to investigate how AI agents with theory of mind capabilities can better anticipate and respond to personality-driven behaviors, potentially improving adaptation strategies in dynamic operational contexts. Finally, we aim to validate our simulation findings through human-in-the-loop experiments, ensuring that our framework translates effectively from simulated to real-world environments.

\newpage
\bibliographystyle{unsrt}
\bibliography{references,references_trust}

\begin{thebibliography}{10}

\bibitem{zhouSOTOPIAInteractiveEvaluation2024}
Xuhui Zhou, Hao Zhu, Leena Mathur, Ruohong Zhang, Haofei Yu, Zhengyang Qi, Louis-Philippe Morency, Yonatan Bisk, Daniel Fried, Graham Neubig, and Maarten Sap.
\newblock {{SOTOPIA}}: {{Interactive Evaluation}} for {{Social Intelligence}} in {{Language Agents}}, March 2024.

\bibitem{volkovaVirTLabAugmentedIntelligenceinpress}
Svitlana Volkova, Daniel Nguyen, Hsien-Te Kao, Myke~C. Cohen, Grant Engberson, Laura Cassani, Trenton~W. Ford, Michael~G. Yankoski, Mohammed Almutairi, Charles Chiang, Nandini Banerjee, Matthew Belcher, Tim Weninger, and Diego {Gomez-Zara}.
\newblock {{VirTLab}}: {{Augmented Intelligence}} for {{Modeling}} and {{Evaluating Human-AI Teaming}} through {{Agent Interactions}}.
\newblock in press.

\bibitem{nguyenExploratoryModelsHumanAI2024}
Daniel Nguyen, Myke~C. Cohen, Hsien-Te Kao, Grant Engberson, Louis Penafiel, Spencer Lynch, and Svitlana Volkova.
\newblock Exploratory {{Models}} of {{Human-AI Teams}}: {{Leveraging Human Digital Twins}} to {{Investigate Trust Development}}, November 2024.

\bibitem{cuiHumanAutonomyTeamingAutonomous2023}
Can Cui, Yunsheng Ma, Xu~Cao, Wenqian Ye, and Ziran Wang.
\newblock Human-{{Autonomy Teaming}} on {{Autonomous Vehicles}} with {{Large Language Model-Enabled Human Digital Twins}}.
\newblock In {\em 2023 {{IEEE}}/{{ACM Symposium}} on {{Edge Computing}} ({{SEC}})}, pages 319--324, December 2023.

\bibitem{yangLLMBasedDigitalTwin2024}
Hanqing Yang, Marie Siew, and Carlee {Joe-Wong}.
\newblock An {{LLM-Based Digital Twin}} for {{Optimizing Human-in-the Loop Systems}}, March 2024.

\bibitem{bendellIndividualTeamProfiling2024}
Rhyse Bendell, Jessica Williams, Stephen~M. Fiore, and Florian Jentsch.
\newblock Individual and team profiling to support theory of mind in artificial social intelligence.
\newblock {\em Scientific Reports}, 14(1):12635, June 2024.

\bibitem{furnhamPersonalityInterpersonalInfluence2024}
Adrian Furnham, Stephen Cuppello, and David~S. Semmelink.
\newblock Personality and {{Interpersonal Influence}}: {{Low Adjustment}} and {{Low Competitiveness}} is {{Associated With Low Assertiveness}}.
\newblock {\em Psychological Reports}, page 00332941241246201, November 2024.

\bibitem{huangHowPersonalityTraits2024}
Yin~Jou Huang and Rafik Hadfi.
\newblock How {{Personality Traits Influence Negotiation Outcomes}}? {{A Simulation}} based on {{Large Language Models}}.
\newblock In Yaser {Al-Onaizan}, Mohit Bansal, and Yun-Nung Chen, editors, {\em Findings of the {{Association}} for {{Computational Linguistics}}: {{EMNLP}} 2024}, pages 10336--10351, Miami, Florida, USA, November 2024. Association for Computational Linguistics.

\bibitem{shadishExperimentalQuasiExperimentalDesigns2001a}
William~R. Shadish, Thomas~D. Cook, and Donald~T. Campbell.
\newblock {\em Experimental and {{Quasi-Experimental Designs}} for {{Generalized Causal Inference}}}.
\newblock Cengage Learning, Belmont, CA, 2nd edition edition, January 2001.

\bibitem{beaumontCausalNex2021}
Paul Beaumont, Ben Horsburgh, Philip Pilgerstorfer, Angel Droth, Richard Oentaryo, Steven Ler, Hiep Nguyen, Gabriel~Azevedo Ferreira, Zain Patel, and Wesley Leong.
\newblock {{CausalNex}}, October 2021.

\bibitem{atheyGeneralizedRandomForests2019b}
Susan Athey, Julie Tibshirani, and Stefan Wager.
\newblock Generalized random forests.
\newblock {\em The Annals of Statistics}, 47(2):1148--1178, April 2019.

\bibitem{mccraeIntroductionFiveFactorModel1992}
Robert~R. McCrae and Oliver~P. John.
\newblock An {{Introduction}} to the {{Five}}-{{Factor Model}} and {{Its Applications}}.
\newblock {\em Journal of Personality}, 60(2):175--215, June 1992.

\bibitem{allport1936trait}
Gordon~W Allport and Henry~S Odbert.
\newblock Trait-names: {{A}} psycho-lexical study.
\newblock {\em Psychological monographs}, 47(1):i, 1936.

\bibitem{cattell1946description}
Raymond~Bernard Cattell.
\newblock Description and measurement of personality.
\newblock 1946.

\bibitem{fiske1949consistency}
Donald~W Fiske.
\newblock Consistency of the factorial structures of personality ratings from different sources.
\newblock {\em The Journal of Abnormal and Social Psychology}, 44(3):329, 1949.

\bibitem{tupes1961recurrent}
E.~C. Tupes and R.~E. Christal.
\newblock Recurrent personality factors based on trait ratings.
\newblock {{USAF ASD Tech}}. {{Rep}}. {{No}}. 61-97, US Air Force, Lackland Air Force Base, TX, 1961.

\bibitem{norman1963toward}
Warren~T Norman.
\newblock Toward an adequate taxonomy of personality attributes: Replicated factor structure in peer nomination personality ratings.
\newblock {\em The journal of abnormal and social psychology}, 66(6):574, 1963.

\bibitem{costa2008revised}
Paul~T Costa and Robert~R McCrae.
\newblock The revised neo personality inventory (neo-pi-r).
\newblock {\em The SAGE handbook of personality theory and assessment}, 2(2):179--198, 2008.

\bibitem{frischLLMAgentsInteraction2024}
Ivar Frisch and Mario Giulianelli.
\newblock {{LLM Agents}} in {{Interaction}}: {{Measuring Personality Consistency}} and {{Linguistic Alignment}} in {{Interacting Populations}} of {{Large Language Models}}, February 2024.

\bibitem{karraEstimatingPersonalityWhiteBox2023}
Saketh~Reddy Karra, Son~The Nguyen, and Theja Tulabandhula.
\newblock Estimating the {{Personality}} of {{White-Box Language Models}}, May 2023.

\bibitem{huangWhoChatGPTBenchmarking2024}
Jen-tse Huang, Wenxuan Wang, Eric~John Li, Man~Ho Lam, Shujie Ren, Youliang Yuan, Wenxiang Jiao, Zhaopeng Tu, and Michael~R. Lyu.
\newblock Who is {{ChatGPT}}? {{Benchmarking LLMs}}' {{Psychological Portrayal Using PsychoBench}}, January 2024.

\bibitem{jiangPersonaLLMInvestigatingAbility2024}
Hang Jiang, Xiajie Zhang, Xubo Cao, Cynthia Breazeal, Deb Roy, and Jad Kabbara.
\newblock {{PersonaLLM}}: {{Investigating}} the {{Ability}} of {{Large Language Models}} to {{Express Personality Traits}}, April 2024.

\bibitem{duanPowerPersonalityHuman2025}
Yifan Duan, Yihong Tang, Xuefeng Bai, Kehai Chen, Juntao Li, and Min Zhang.
\newblock The {{Power}} of {{Personality}}: {{A Human Simulation Perspective}} to {{Investigate Large Language Model Agents}}, February 2025.

\bibitem{aiSelfknowledgeActionConsistent2024}
Yiming Ai, Zhiwei He, Ziyin Zhang, Wenhong Zhu, Hongkun Hao, Kai Yu, Lingjun Chen, and Rui Wang.
\newblock Is {{Self-knowledge}} and {{Action Consistent}} or {{Not}}: {{Investigating Large Language Model}}'s {{Personality}}, December 2024.

\bibitem{petrovLimitedAbilityLLMs2024}
Nikolay~B. Petrov, Gregory {Serapio-Garc{\'i}a}, and Jason Rentfrow.
\newblock Limited {{Ability}} of {{LLMs}} to {{Simulate Human Psychological Behaviours}}: A {{Psychometric Analysis}}, May 2024.

\bibitem{molchanovaExploringPotentialLarge2025}
Maria Molchanova, Anna Mikhailova, Anna Korzanova, Lidiia Ostyakova, and Alexandra Dolidze.
\newblock Exploring the {{Potential}} of {{Large Language Models}} to {{Simulate Personality}}, February 2025.

\bibitem{raiffaArtScienceNegotiation1982}
Howard Raiffa.
\newblock {\em The {{Art}} and {{Science}} of {{Negotiation}}}.
\newblock Harvard University Press, 1982.

\bibitem{gilkey1986role}
Roderick~W Gilkey and Leonard Greenhalgh.
\newblock The role of personality in successful negotiating.
\newblock {\em Negotiation Journal}, 2(3):245--256, 1986.

\bibitem{sassPersonalityNegotiationPerformance2015}
Mary Sass and Matthew {Liao-Troth}.
\newblock Personality and {{Negotiation Performance}}: {{The People Matter}}, January 2015.

\bibitem{kangHowInterviewersRespond2015}
Gerui~(Grace) Kang, Lin Xiu, and Alan~C. Roline.
\newblock How do interviewers respond to applicants' initiation of salary negotiation? {{An}} exploratory study on the role of gender and personality.
\newblock {\em Evidence-based HRM: a Global Forum for Empirical Scholarship}, 3(2):145--158, August 2015.

\bibitem{barry1998bargainer}
Bruce Barry and Raymond~A Friedman.
\newblock Bargainer characteristics in distributive and integrative negotiation.
\newblock {\em Journal of personality and social psychology}, 74(2):345, 1998.

\bibitem{amanatullahNegotiatorsWhoGive2008}
Emily~T. Amanatullah, Michael~W. Morris, and Jared~R. Curhan.
\newblock Negotiators who give too much: {{Unmitigated}} communion, relational anxieties, and economic costs in distributive and integrative bargaining.
\newblock {\em Journal of Personality and Social Psychology}, 95(3):723--738, 2008.

\bibitem{gormanTeamCoordinationDynamics2010}
Jamie~C Gorman, Polemnia~G Amazeen, and Nancy~J Cooke.
\newblock Team coordination dynamics.
\newblock {\em Nonlinear dynamics, psychology, and life sciences}, 14(3):265--289, July 2010.

\bibitem{demirTeamCoordinationDynamics2017}
Mustafa Demir, Polomnia~G. Amazeen, Nathan~J. McNeese, Aaron Likens, and Nancy~J. Cooke.
\newblock Team {{Coordination Dynamics}} in {{Human-Autonomy Teaming}}.
\newblock {\em Proceedings of the Human Factors and Ergonomics Society Annual Meeting}, 61(1):236--236, September 2017.

\bibitem{falcaoBigFivePersonality2018}
Pedro~Fontes Falc{\~a}o, Manuel Saraiva, Eduardo Santos, and Miguel Pina~E Cunha.
\newblock Big {{Five}} personality traits in simulated negotiation settings.
\newblock {\em EuroMed Journal of Business}, 13(2):201--213, July 2018.

\bibitem{pennebakerLinguisticStylesLanguage1999}
James~W. Pennebaker and Laura~A. King.
\newblock Linguistic styles: {{Language}} use as an individual difference.
\newblock {\em Journal of Personality and Social Psychology}, 77(6):1296--1312, 1999.

\bibitem{tausczikPsychologicalMeaningWords2010}
Yla~R. Tausczik and James~W. Pennebaker.
\newblock The {{Psychological Meaning}} of {{Words}}: {{LIWC}} and {{Computerized Text Analysis Methods}}.
\newblock {\em Journal of Language and Social Psychology}, 29(1):24--54, March 2010.

\bibitem{chawla2023towards}
Kushal Chawla, Rene Clever, Jaysa Ramirez, Gale~M Lucas, and Jonathan Gratch.
\newblock Towards emotion-aware agents for improved user satisfaction and partner perception in negotiation dialogues.
\newblock {\em IEEE Transactions on Affective Computing}, 2023.

\bibitem{grazianoAgreeablenessEmpathyHelping2007}
William~G. Graziano, Meara~M. Habashi, Brad~E. Sheese, and Ren{\'e}e~M. Tobin.
\newblock Agreeableness, empathy, and helping: {{A}} person {\texttimes} situation perspective.
\newblock {\em Journal of Personality and Social Psychology}, 93(4):583--599, 2007.

\bibitem{leeDoesGPT3Generate2022}
Young-Jun Lee, Chae-Gyun Lim, and Ho-Jin Choi.
\newblock Does {{GPT-3 Generate Empathetic Dialogues}}? {{A Novel In-Context Example Selection Method}} and {{Automatic Evaluation Metric}} for {{Empathetic Dialogue Generation}}.
\newblock In Nicoletta Calzolari, Chu-Ren Huang, Hansaem Kim, James Pustejovsky, Leo Wanner, Key-Sun Choi, Pum-Mo Ryu, Hsin-Hsi Chen, Lucia Donatelli, Heng Ji, Sadao Kurohashi, Patrizia Paggio, Nianwen Xue, Seokhwan Kim, Younggyun Hahm, Zhong He, Tony~Kyungil Lee, Enrico Santus, Francis Bond, and Seung-Hoon Na, editors, {\em Proceedings of the 29th {{International Conference}} on {{Computational Linguistics}}}, pages 669--683, Gyeongju, Republic of Korea, October 2022. International Committee on Computational Linguistics.

\bibitem{garten2016morality}
Justin Garten, Reihane Boghrati, Joe Hoover, Kate~M Johnson, and Morteza Dehghani.
\newblock Morality between the lines: {{Detecting}} moral sentiment in text.
\newblock In {\em Proceedings of {{IJCAI}} 2016 Workshop on Computational Modeling of Attitudes}, 2016.

\bibitem{rashkinConnotationFramesDataDriven2016a}
Hannah Rashkin, Sameer Singh, and Yejin Choi.
\newblock Connotation {{Frames}}: {{A Data-Driven Investigation}}, August 2016.

\bibitem{grahamLiberalsConservativesRely2009}
Jesse Graham, Jonathan Haidt, and Brian~A. Nosek.
\newblock Liberals and conservatives rely on different sets of moral foundations.
\newblock {\em Journal of Personality and Social Psychology}, 96(5):1029--1046, May 2009.

\bibitem{hanuDetoxify2020}
Laura Hanu and {Unitary team}.
\newblock Detoxify, November 2020.

\bibitem{curhanWhatPeopleValue2006}
Jared~R. Curhan, Hillary~Anger Elfenbein, and Heng Xu.
\newblock What do people value when they negotiate? {{Mapping}} the domain of subjective value in negotiation.
\newblock {\em Journal of Personality and Social Psychology}, 91(3):493--512, September 2006.

\bibitem{elfenbeinAreNegotiatorsBetter2008}
Hillary~Anger Elfenbein, Jared~R. Curhan, Noah Eisenkraft, Aiwa Shirako, and Lucio Baccaro.
\newblock Are {{Some Negotiators Better Than Others}}? {{Individual Differences}} in {{Bargaining Outcomes}}.
\newblock {\em Journal of research in personality}, 42(6):1463--1475, December 2008.

\bibitem{dimotakisMindHeartLiterally2012}
Nikolaos Dimotakis, Donald~E. Conlon, and Remus Ilies.
\newblock The mind and heart (literally) of the negotiator: {{Personality}} and contextual determinants of experiential reactions and economic outcomes in negotiation.
\newblock {\em Journal of Applied Psychology}, 97(1):183--193, 2012.

\bibitem{prajodEffectVirtualAgent2019}
Pooja Prajod, Mohammed~Al Owayyed, and Tim Rietveld.
\newblock The {{Effect}} of {{Virtual Agent Warmth}} on {{Human-Agent Negotiation}}.
\newblock 2019.

\bibitem{zhouTrustingVirtualAgents2019}
Michelle~X. Zhou, Gloria Mark, Jingyi Li, and Huahai Yang.
\newblock Trusting {{Virtual Agents}}: {{The Effect}} of {{Personality}}.
\newblock {\em ACM Trans. Interact. Intell. Syst.}, 9(2-3):10:1--10:36, March 2019.

\bibitem{Zhou2024SOTOPIA}
Xuhui Zhou, Hao Zhu, Leena Mathur, Ruohong Zhang, Zhengyang Qi, Haofei Yu, Louis-Philippe Morency, Yonatan Bisk, Daniel Fried, Graham Neubig, and Maarten Sap.
\newblock Sotopia: Interactive evaluation for social intelligence in language agents.
\newblock {\em International Conference on Learning Representations (ICLR)}, 2024.

\bibitem{see2019makes}
Abigail See, Stephen Roller, Douwe Kiela, and Jason Weston.
\newblock What makes a good conversation? how controllable attributes affect human judgments.
\newblock {\em arXiv preprint arXiv:1902.08654}, 2019.

\bibitem{rashkinConnotationFramesDataDriven2016}
Hannah Rashkin, Sameer Singh, and Yejin Choi.
\newblock Connotation {{Frames}}: {{A Data-Driven Investigation}}, August 2016.

\bibitem{graham2013moral}
Jesse Graham, Jonathan Haidt, Sena Koleva, Matt Motyl, Ravi Iyer, Sean~P Wojcik, and Peter~H Ditto.
\newblock Moral foundations theory: The pragmatic validity of moral pluralism.
\newblock In {\em Advances in experimental social psychology}, volume~47, pages 55--130. Elsevier, 2013.

\bibitem{rashkinTruthVaryingShades2017}
Hannah Rashkin, Eunsol Choi, Jin~Yea Jang, Svitlana Volkova, and Yejin Choi.
\newblock Truth of {{Varying Shades}}: {{Analyzing Language}} in {{Fake News}} and {{Political Fact-Checking}}.
\newblock In Martha Palmer, Rebecca Hwa, and Sebastian Riedel, editors, {\em Proceedings of the 2017 {{Conference}} on {{Empirical Methods}} in {{Natural Language Processing}}}, pages 2931--2937, Copenhagen, Denmark, September 2017. Association for Computational Linguistics.

\bibitem{sanhDistilBERTDistilledVersion2020}
Victor Sanh, Lysandre Debut, Julien Chaumond, and Thomas Wolf.
\newblock {{DistilBERT}}, a distilled version of {{BERT}}: Smaller, faster, cheaper and lighter, March 2020.

\bibitem{Detoxify}
Laura Hanu and {Unitary team}.
\newblock Detoxify.
\newblock Github. https://github.com/unitaryai/detoxify, 2020.

\bibitem{savaniDistilBERTEmotionRecognition2024}
Bhadresh Savani.
\newblock {{DistilBERT}} for emotion recognition, May 2024.

\bibitem{volkova2021machine}
M~Glenski, E~Ayton, E~Saldanha, J~Mendoza, D~Arendt, Z~Shaw, K~Cronk, S~Smith, and M~Greaves.
\newblock Machine intelligence to detect, characterise, and defend against influence operations in the information environment.
\newblock {\em Journal of Information Warfare}, 20(2):42--66, 2021.

\bibitem{devlinBERTPretrainingDeep2019}
Jacob Devlin, Ming-Wei Chang, Kenton Lee, and Kristina Toutanova.
\newblock {{BERT}}: {{Pre-training}} of {{Deep Bidirectional Transformers}} for {{Language Understanding}}, May 2019.

\bibitem{aluruDeepDiveMultilingual2021}
Sai~Saketh Aluru, Binny Mathew, Punyajoy Saha, and Animesh Mukherjee.
\newblock A {{Deep Dive}} into {{Multilingual Hate Speech Classification}}.
\newblock In Yuxiao Dong, Georgiana Ifrim, Dunja Mladeni{\'c}, Craig Saunders, and Sofie Van~Hoecke, editors, {\em Machine {{Learning}} and {{Knowledge Discovery}} in {{Databases}}. {{Applied Data Science}} and {{Demo Track}}}, pages 423--439, Cham, 2021. Springer International Publishing.

\bibitem{he2018decoupling}
He~He, Derek Chen, Anusha Balakrishnan, and Percy Liang.
\newblock Decoupling strategy and generation in negotiation dialogues, 2018.

\bibitem{pearlBookWhyNew2018}
Judea Pearl and Dana Mackenzie.
\newblock {\em The {{Book}} of {{Why}}: {{The New Science}} of {{Cause}} and {{Effect}}}.
\newblock Basic Books, New York, 1st edition edition, May 2018.

\bibitem{battocchiEconMLPythonPackage2019}
Keith Battocchi, Eleanor Dillon, Maggie Hei, Greg Lewis, Paul Oka, Miruna Oprescu, and Vasilis Syrgkanis.
\newblock {EconML}: A python package for ml-based heterogeneous treatment effects estimation.
\newblock \url{https://github.com/py-why/EconML}, 2019.
\newblock Version 0.x.

\bibitem{bradleyTeamPlayersCollective2013}
Bret~H. Bradley, John~E. Baur, Christopher~G. Banford, and Bennett~E. Postlethwaite.
\newblock Team {{Players}} and {{Collective Performance}}: {{How Agreeableness Affects Team Performance Over Time}}.
\newblock {\em Small Group Research}, 44(6):680--711, December 2013.

\bibitem{driskellWhatMakesGood2006}
James~E. Driskell, Gerald~F. Goodwin, Eduardo Salas, and Patrick~Gavan O'Shea.
\newblock What makes a good team player? {{Personality}} and team effectiveness.
\newblock {\em Group Dynamics: Theory, Research, and Practice}, 10(4):249--271, December 2006.

\bibitem{grantRethinkingExtravertedSales2013}
Adam~M. Grant.
\newblock Rethinking the {{Extraverted Sales Ideal}}: {{The Ambivert Advantage}}.
\newblock {\em Psychological Science}, 24(6):1024--1030, June 2013.

\bibitem{lepineAdaptabilityChangingTask2000}
Jeffrey~A. Lepine, Jason~A. Colquitt, and Amir Erez.
\newblock Adaptability to {{Changing Task Contexts}}: {{Effects}} of {{General Cognitive Ability}}, {{Conscientiousness}}, and {{Openness}} to {{Experience}}.
\newblock {\em Personnel Psychology}, 53(3):563--593, 2000.

\bibitem{bellDeeplevelCompositionVariables2007}
Suzanne~T. Bell.
\newblock Deep-level composition variables as predictors of team performance: A meta-analysis.
\newblock {\em The Journal of Applied Psychology}, 92(3):595--615, May 2007.

\bibitem{kleinHOWTHEYGET2004}
K.~J. Klein, J.~L. Saltz, and D.~M. Mayer.
\newblock {{HOW DO THEY GET THERE}}? {{AN EXAMINATION OF THE ANTECEDENTS OF CENTRALITY IN TEAM NETWORKS}}.
\newblock {\em Academy of Management Journal}, 47(6):952--963, December 2004.

\bibitem{peetersPersonalityTeamPerformance2006}
Miranda A.~G. Peeters, Harrie F. J.~M. Van~Tuijl, Christel~G. Rutte, and Isabelle M. M.~J. Reymen.
\newblock Personality and team performance: A meta-analysis.
\newblock {\em European Journal of Personality}, 20(5):377--396, August 2006.

\bibitem{watson1994panas}
David Watson and Lee~Anna Clark.
\newblock The {{PANAS-X}}: {{Manual}} for the positive and negative affect schedule-expanded form.
\newblock 1994.

\bibitem{mccraeNEOPIRData362002}
Robert~R. McCrae.
\newblock {{NEO-PI-R Data}} from 36 {{Cultures}}.
\newblock In Robert~R. McCrae and J{\"u}ri Allik, editors, {\em The {{Five-Factor Model}} of {{Personality Across Cultures}}}, pages 105--125. Springer US, Boston, MA, 2002.

\bibitem{habashiSearchingProsocialPersonality2016}
Meara~M. Habashi, William~G. Graziano, and Ann~E. Hoover.
\newblock Searching for the {{Prosocial Personality}}: {{A Big Five Approach}} to {{Linking Personality}} and {{Prosocial Behavior}}.
\newblock {\em Personality and Social Psychology Bulletin}, 42(9):1177--1192, September 2016.

\bibitem{hirshCompassionateLiberalsPolite2010}
Jacob~B. Hirsh, Colin~G. DeYoung, {Xiaowen Xu}, and Jordan~B. Peterson.
\newblock Compassionate {{Liberals}} and {{Polite Conservatives}}: {{Associations}} of {{Agreeableness With Political Ideology}} and {{Moral Values}}.
\newblock {\em Personality and Social Psychology Bulletin}, 36(5):655--664, May 2010.

\bibitem{watsonExtraversionItsPositive1997}
David Watson and Lee~Anna Clark.
\newblock Extraversion and {{Its Positive Emotional Core}}.
\newblock In {\em Handbook of {{Personality Psychology}}}, pages 767--793. Elsevier, 1997.

\bibitem{hancockEvolvingTrustRobots2021}
P.~A. Hancock, Theresa~T. Kessler, Alexandra~D. Kaplan, John~C. Brill, and James~L. Szalma.
\newblock Evolving {{Trust}} in {{Robots}}: {{Specification Through Sequential}} and {{Comparative Meta-Analyses}}.
\newblock {\em Human Factors}, 63(7):1196--1229, November 2021.

\bibitem{schaeferMetaAnalysisFactorsInfluencing2016}
Kristin~E. Schaefer, Jessie Y.~C. Chen, James~L. Szalma, and P.~A. Hancock.
\newblock A {{Meta-Analysis}} of {{Factors Influencing}} the {{Development}} of {{Trust}} in {{Automation}}: {{Implications}} for {{Understanding Autonomy}} in {{Future Systems}}.
\newblock {\em Human Factors}, 58(3):377--400, May 2016.

\bibitem{hancockMetaAnalysisFactorsAffecting2011}
Peter~A. Hancock, Deborah~R. Billings, Kristin~E. Schaefer, Jessie Y.~C. Chen, Ewart~J. {de Visser}, and Raja Parasuraman.
\newblock A {{Meta-Analysis}} of {{Factors Affecting Trust}} in {{Human-Robot Interaction}}.
\newblock {\em Human Factors: The Journal of the Human Factors and Ergonomics Society}, 53(5):517--527, October 2011.

\bibitem{jessupMeasurementPropensityTrust2019}
Sarah~A. Jessup, Tamera~R. Schneider, Gene~M. Alarcon, Tyler~J. Ryan, and August Capiola.
\newblock The {{Measurement}} of the {{Propensity}} to {{Trust Automation}}.
\newblock In Jessie~Y.C. Chen and Gino Fragomeni, editors, {\em Virtual, {{Augmented}} and {{Mixed Reality}}. {{Applications}} and {{Case Studies}}}, pages 476--489, Cham, 2019. Springer International Publishing.

\end{thebibliography}

\appendix

\appendix

\footnotesize


\newpage

\section{Example Agent Profile}
\label{appendix:agent_profile}
\begin{lstlisting}
"first_name": "Human",
"last_name": "Agent",
"age": 22,
"occupation": "Candidate",
"personality_and_values": Personality Model: Big 5 Personality
Personality Trait: Introversion
Task Assignment: Prefers independent tasks and may struggle with collaboration.
Interaction: Tends to avoid social interactions and may appear distant or reserved.
Communication: May be quiet or withdrawn in communication, leading to misunderstandings.
Planning: Tends to plan independently, potentially missing out on input from others.
Leadership: May prefer to work alone rather than lead a team.
Individual Role: May prefer solitary tasks and independent work."
\end{lstlisting}

\section{Craigslist Scenario Example}
\label{appendix:craigslist}
\begin{lstlisting}
Scenario Description: One person is offering a 47 inch LED TV for a price of $349.0, while another person is showing interest in purchasing it. Here is a description of the TV: This is a stunning 47 inch LED TV in pristine condition. The model is the LG M Series LM476700. The buyer will need to arrange for pick-up in San Ramon. Feel free to call or text if you\'re interested. The TV is smart enabled with WIFI and has built-in apps like Netflix, Amazon, Youtube and more. It comes with a "Magic Remote" that has motion sensor controls. The LED display boasts 1080 HD resolution and also has a 3D function. The design is slim and lightweight with an attractive silver bezel.
Agent 1 Goal: You are the buyer for this item and your target price is \$152.0. You will be penalized if you purchase it at a significantly higher price than the target. However, if you manage to buy it for less than the target price, you'll receive a bonus.
Agent 2 Goal: As the seller of this item, your target price is set at\ $172.5. Please be aware that you will face a penalty if the item is sold for a significantly lower price than the target. However, if you manage to sell it for more than the target price, you will receive a bonus.
\end{lstlisting}

\section{Trait Variation Prompt}
\label{appendix:trait_variation}
\begin{minted}{python}
credibility_persona = {
    "High_Transparency-High_Competence-High_Adaptability": {
        "Task_Assignment": "Delegates tasks with clear explanations, leveraging high competence and adaptability to adjust to evolving needs and challenges.",
        "Interaction": "Engages openly with team members, sharing knowledge and adapting interactions based on feedback and changing circumstances.",
        "Communication": "Communicates transparently and expertly, adapting messages to ensure clarity and relevance for various situations and audiences.",
        "Planning": "Involves the team in detailed, transparent planning processes, with strategies that adapt to new information and changing conditions.",
        "Leadership": "Leads with high transparency and adaptability, using expertise to navigate changes and inspire confidence and flexibility within the team.",
        "Individual_Role": "Known for a high level of openness, skill, and flexibility, significantly contributing to team success by adapting to dynamic environments."
    },
    "High_Transparency-High_Competence-Low_Adaptability": {
        "Task_Assignment": "Assigns tasks with clear and competent guidance but may struggle to adjust plans or strategies in response to unforeseen changes.",
        "Interaction": "Maintains open communication and provides expert input, though may not easily adapt interactions to rapidly changing team dynamics or feedback.",
        "Communication": "Communicates effectively and transparently, but may find it challenging to modify communication styles or approaches as situations evolve.",
        "Planning": "Creates detailed plans with clear transparency and high competence, but may have difficulty adapting strategies if new information or changes arise.",
        "Leadership": "Leads with clarity and expertise, though adaptability might be limited, potentially affecting the ability to respond effectively to unexpected changes.",
        "Individual_Role": "Provides high-quality and transparent input but may need to improve flexibility to better handle evolving situations."
    },
    "High_Transparency-Low_Competence-High_Adaptability": {
        "Task_Assignment": "Delegates tasks with openness and clarity but may lack the expertise needed for effective execution, while adapting to team needs and feedback.",
        "Interaction": "Engages openly with team members, adapting interactions based on feedback, though might not offer deep or technically sound guidance due to lower competence.",
        "Communication": "Communicates transparently and adjusts messaging based on context and feedback, though may lack depth and technical detail in explanations.",
        "Planning": "Shares planning processes openly and adapts strategies based on new information, though plans may lack the necessary competence for optimal execution.",
        "Leadership": "Promotes transparency and flexibility but may struggle with providing expert guidance, requiring continuous adaptation to improve effectiveness.",
        "Individual_Role": "Creates an open and adaptable environment but needs to bolster competence to enhance overall effectiveness and contribution."
    },
    "High_Transparency-Low_Competence-Low_Adaptability": {
        "Task_Assignment": "Assigns tasks with clear instructions but struggles with effective execution due to low competence and adaptability, providing minimal updates.",
        "Interaction": "Interacts transparently but may be rigid and less responsive to feedback or changing conditions, impacting support and team dynamics.",
        "Communication": "Communicates clearly but may lack depth and flexibility, leading to incomplete or inadequate guidance due to limited expertise and adaptability.",
        "Planning": "Shares planning details openly but with limited effectiveness and adaptability, resulting in suboptimal strategies and execution challenges.",
        "Leadership": "Demonstrates transparency but struggles with both competence and adaptability, leading to less effective leadership and team direction.",
        "Individual_Role": "Known for clear but ineffective communication and lack of adaptability, requiring significant improvement in skill and flexibility for effective contribution."
    },
    "Low_Transparency-High_Competence-High_Adaptability": {
        "Task_Assignment": "Delegates tasks effectively based on high competence and adaptability but with limited transparency in updates or rationale.",
        "Interaction": "Engages positively with team members while adapting interactions based on changing needs, though may not share all relevant information.",
        "Communication": "Provides knowledgeable input and adjusts communication style as needed, though might not be fully transparent about processes or details.",
        "Planning": "Develops effective and adaptable plans but keeps details and rationale guarded, potentially impacting overall team alignment and understanding.",
        "Leadership": "Leads with strong skill and adaptability but maintains some level of secrecy, affecting team trust and cohesion despite effective execution.",
        "Individual_Role": "Demonstrates high competence and flexibility but may need to increase transparency to enhance overall team effectiveness and collaboration."
    },
    "Low_Transparency-High_Competence-Low_Adaptability": {
        "Task_Assignment": "Assigns tasks with high competence but limited transparency and adaptability, resulting in unclear guidance and difficulty responding to changes.",
        "Interaction": "Interacts with caution and minimal openness, providing skilled support but struggling to adapt interactions based on team feedback or changes.",
        "Communication": "Communicates authoritatively but with limited transparency, and may struggle to adjust messages based on evolving needs or contexts.",
        "Planning": "Creates detailed plans with high expertise but lacks adaptability and transparency, leading to potential gaps in team understanding and responsiveness.",
        "Leadership": "Leads with high skill but limited adaptability and openness, which may impact team cohesion and effectiveness despite competent execution.",
        "Individual_Role": "Known for high competence but requires improvement in transparency and adaptability to fully support team dynamics and responsiveness."
    },
    "Low_Transparency-Low_Competence-High_Adaptability": {
        "Task_Assignment": "Delegates tasks with minimal competence and transparency but shows high adaptability in adjusting approaches based on team feedback and changes.",
        "Interaction": "Engages with team members in a flexible manner but may lack depth in technical guidance and provide limited information.",
        "Communication": "Communicates with adaptability but limited clarity and expertise, leading to potential misunderstandings and ineffective guidance.",
        "Planning": "Plans with high adaptability but minimal transparency and competence, resulting in unclear and potentially ineffective strategies.",
        "Leadership": "Demonstrates flexibility and responsiveness but struggles with both transparency and skill, affecting overall leadership effectiveness.",
        "Individual_Role": "Creates an adaptable environment but requires significant improvement in competence and transparency to enhance overall effectiveness."
    },
    "Low_Transparency-Low_Competence-Low_Adaptability": {
        "Task_Assignment": "Assigns tasks with reluctance and minimal effectiveness, lacking competence, transparency, and adaptability, resulting in poor outcomes.",
        "Interaction": "Interacts in a guarded manner with limited information sharing and adaptability, providing minimal support and demonstrating low skill.",
        "Communication": "Shares minimal and unclear information, leading to confusion and ineffective communication within the team due to low competence and flexibility.",
        "Planning": "Plans with minimal effectiveness and adaptability, resulting in unclear strategies and challenges in execution due to low competence and transparency.",
        "Leadership": "Struggles with leadership due to low trust, transparency, competence, and adaptability, leading to poor team dynamics and performance.",
        "Individual_Role": "Considered ineffective and uncommunicative, requiring substantial improvement across transparency, competence, and adaptability."
    }
}
\end{minted}

\section{Job Negotiation details}
\label{appendix:job_negotiation}
Here we provide the detailed setting of Human-AI Job Negotiation in Section \ref{sec:exp2}. Table \ref{tab:scenario_comparison} shows the score allocations on different choices for two roles.

\begin{table}[h!]
\centering
\setlength{\tabcolsep}{4pt}
\begin{tabular}{@{}lccccc@{}}
\toprule
\textbf{Starting Date} & 6.1 & 6.15 & 7.1 & 7.15 & 8.1 \\ \midrule
Manager       & 0      & 600     & 1200   & 1800    & 2400    \\
Candidate       & 2400   & 1800    & 1200   & 600     & 0       \\ \midrule
\textbf{Salary (\$k)}        & 100 &  105 & 110 & 115 & 120 \\ \midrule
Manager      & 6000   & 4500    & 3000   & 1500    & 0       \\
Candidate       & 0      & 1500    & 3000   & 4500    & 6000    \\ \bottomrule
\end{tabular}
\caption{Comparison of Scenarios for Starting Date and Salary (Candidate vs. Recruiter Points) 
}
\label{tab:scenario_comparison}
\end{table}

\end{document}